# Computing on the Fly

## Navigating a Vision for the Future of Drone Computing

**July 2026**

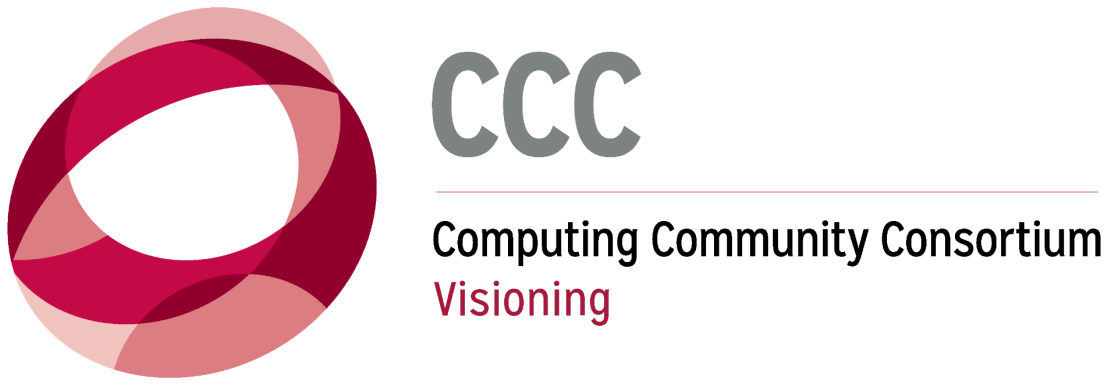

## Authors

Kevin Butler, University of Florida
Christopher Stewart, The Ohio State University
Nils Aschenbruck, Osnabrück University
Alina Gerall, Computing Research Association
Weisong Shi, University of Delaware
Ufuk Topcu, University of Texas
Deborah Silver, Rutgers University

### With Contributions from:

Sabine Brunswicker, Purdue University
Mike Kronander, Burgess & Niple, Inc.
Srinivasan Parthasarathy, The Ohio State University
Shashi Shekhar, University of Minnesota
Quinn Spadola, Computing Research Association
Andy Thurling, DroneUp

## Suggested Citation

## About the Computing Community Consortium (CCC)

A programmatic committee of the Computing Research Association (CRA), CCC enables the pursuit of innovative, high-impact computing research that aligns with pressing national and global challenges. Of, by, and for the computing research community, CCC is a responsive, respected, and visionary organization that brings together thought leaders from industry, academia, and government to articulate and advance compelling research visions and communicate them to stakeholders, policymakers, the public, and the broad computing research community.

# TABLE OF CONTENTS

## 1. Executive Summary

In December 2025, the Computing Community Consortium's (CCC) Computing on the Fly: Navigating a Vision for the Future of Drone Computing workshop convened 47 experts from academia, industry, and government to assess the potential of AI-powered drones for non-military use by 2035 and identify critical barriers to realizing that future. Collectively, the attendees agreed that greater computational support for drones (i.e., computing on the fly) could unlock transformative applications, from wildfire detection and suppression to reliable rural medical supply chains to AI-driven public safety response. However, to realize this future, core research challenges in computing, networking, security, AI, and policy must be addressed. Specifically, we identified the following 12 technical challenges:

**12 Technical Challenges**

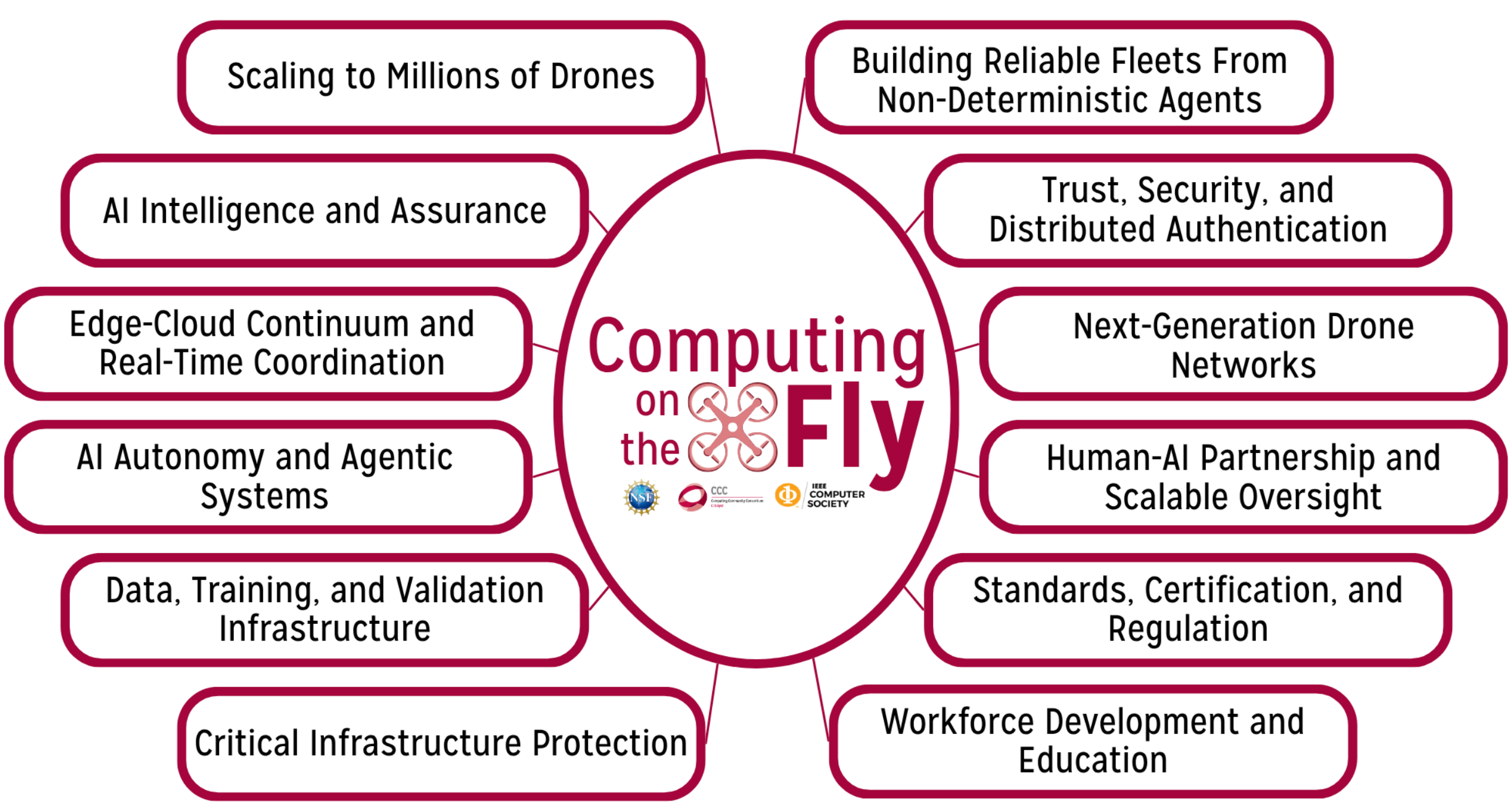

These challenges must be addressed in a timely fashion to realize the transformative potential of this technology. The following catalyzing recommendations represent key checkpoints to ensure timely progress toward this future.

- **Create simulation-to-reality pipelines** using digital twin technologies and hybrid simulation approaches combining synthetic and real-world data to validate million-drone behaviors before physical deployment, bridging the sim-to-real gap.
- **Establish integration testbeds** (physical + digital) supporting graduated complexity, realistic failure injection, and shared milestones for reproducible research.

- **Establish scalable and performant drone-based entity tags and key infrastructure** for identification and tracking at device level.
- **Develop real-time-aware autonomy frameworks** exposing explicit timing, energy, and Quality-of-Service interfaces enabling analyzable composition and runtime adaptation, with standardized cross-layer reconfiguration APIs allowing drones and edge nodes to negotiate sensing rates, model fidelity, and control frequencies under enforced safety constraints and predictable timing guarantees.

These milestones motivate immediate, actionable steps to address the challenges:

| | |
|---|---|
| **Funders** | Coordinate federal investment (i.e., NSF, USDA, DOT/FAA, FCC, and DOD). Establish a national testbed consortium. Target first 1,000-drone demonstration by 2028. |
| **Researchers** | Break silos across robotics, networking, AI, and security. Engage real deployments. Share datasets. Build interdisciplinary graduate programs. |
| **Policymakers** | Adopt open standards (40–60% integration cost reduction potential). Participate in testbeds. Share failure-mode learnings. Fund workforce development efforts. |
| **Industry** | Proactive frameworks across Actor, System, and Infrastructure pillars. Launch regulatory sandboxes. Coordinate internationally to preserve U.S. leadership. |

## Computing on the Fly: Recommendation Topics

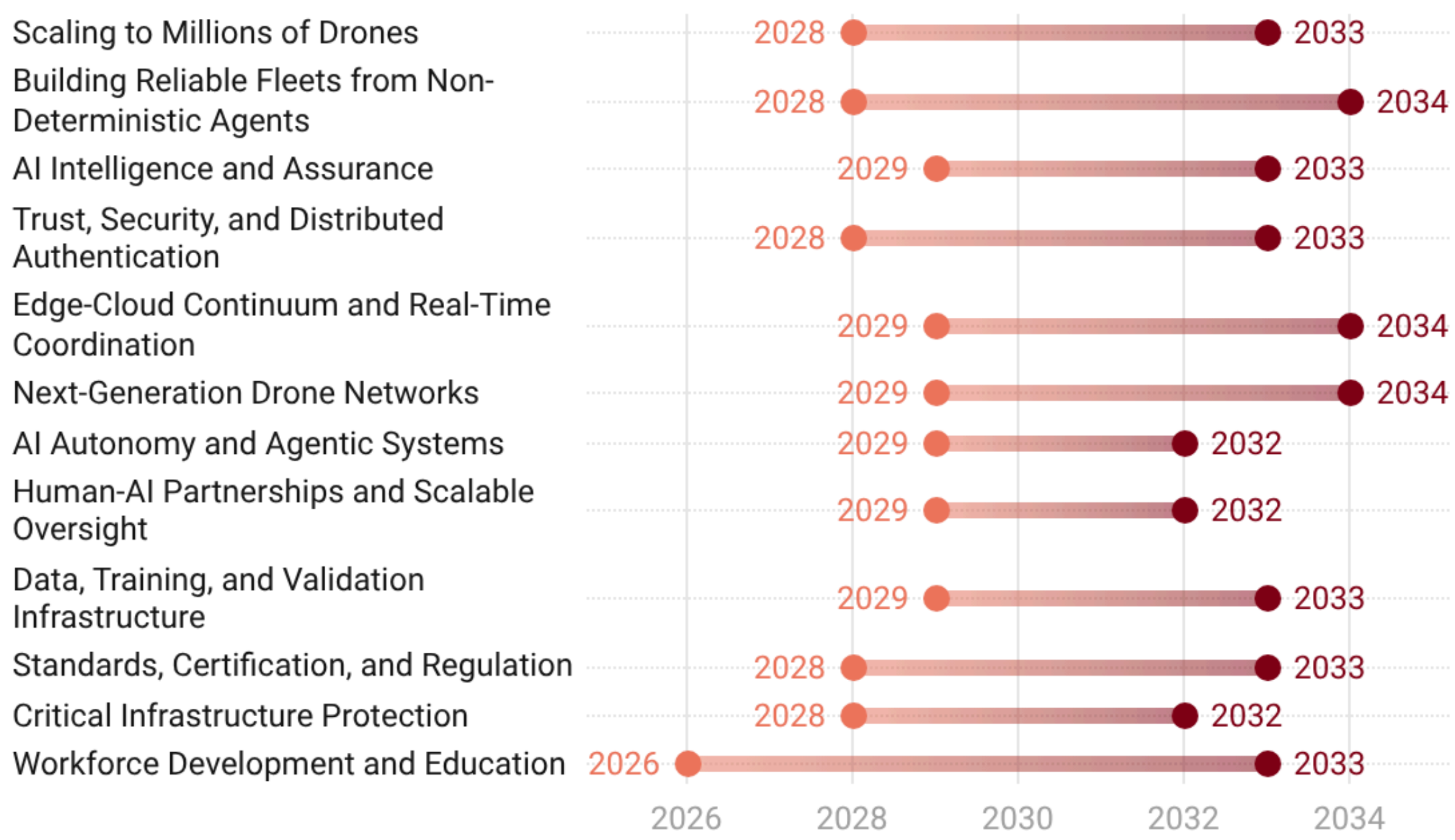

## 2. Vision and Motivation

By 2035, the capabilities, economic opportunities, and societal benefits of AI-powered drones[1] for non-military use will be evident. Drones will move goods, medical supplies, information, and even people, creating an opportunity for infrastructure investment that will be comparable to highways and the electric grid. Imagine the potential impact of the following drone applications:

- **Natural disaster detection drones** that will monitor vast areas and spot smoldering ignition sources within minutes, cutting annual fire suppression costs by billions and protecting watersheds that serve 77 million Americans (United States Forest Service, 2024).
- **Medical supply chains** that will bypass ground congestion, delivering time-sensitive treatments to rural hospitals and disaster zones with 99.9% reliability (Peters, 2026).
- **Fleets of nationwide infrastructure inspection drones** that will monitor 615,000 bridges and 5.5 million miles of power lines, shifting manual, multi-year cycles to continuous, automated assessment (American Society of Civil Engineers, 2025).
- **AI-driven drones that respond to emergencies with unprecedented speed** will enable public-safety officials to more rapidly address emergencies (Skydio, n.d.).
- **Drones will become ubiquitous in agriculture** by safely coordinating local flights to autonomously capture data across multiple modalities and spectra, boosting crop yields, and producing recommendations that help farmers generate predictable profits.

Three trends are driving the rapid trajectory toward advanced drone applications. The first trend is technological maturation. Weight and power demands for drones of all sizes have come down. The energy density of lithium-sulfur batteries has improved greatly in recent years, processors running on devices deployed at the network edge (away from datacenters) can now run AI efficiently, and 6G research promises sub-millisecond latency and micro-second synchronization. Collectively, commercial off-the-shelf drones can now reliably conduct missions that involve complex computation, sensing, and communication.

This has spurred the second trend: policy and market velocity. The FAA projects 2.6 million commercial drones by 2035 — triple today's numbers — with Beyond-Visual-Line-Of-Sight (BVLOS) operations becoming routine by 2027 (Federal Aviation Administration, n.d.). The Department of Defense's "Replicator" initiative aims to field thousands of autonomous systems within 24 months to address critical infrastructure protection (Department of Defense, 2024). The global drone services market's compound annual growth rate is 22%, and is expected to reach $128B by 2030 (Fortune Business Insights, 2026).

[1] For definitions of technical terms, please reference the glossary at the end of this report

Finally, the third trend driving this zeitgeist is growing optimism for national imperatives that can be addressed by physical intelligence embodied in edge computing systems. Labor shortages in aviation already exceed 5% and are expected to grow (United States Government Accountability Office, 2023), weather-caused disasters cost $150B annually (National Oceanic and Atmospheric Administration, n.d.), and research prototypes in digital agriculture and biodiversity and other early use-inspired scenarios show great promise despite relatively low investment so far.

While significant growth in advanced drone applications is inevitable, there remain significant unresolved computational challenges. For example, can we trust the behavior of many coordinated drones in the presence of adversarial agents? Can drones share and leverage edge computing, cloud computing, and mobile computing resources to make advanced drone applications affordable, efficient, and efficacious? Can the U.S. create an environment that facilitates testing and scaling drone applications from ideation to robust, product-ready systems? If these challenges are not addressed, they will act as headwinds, limiting the transformative impact of this unique convergence of synergistic trends.

Today, we face a “capability gap” where hardware and aspirations are outpacing the software and systems needed to operate safely at scale. In December 2025, we convened the Computing On The Fly workshop, including 47 experts from academia, industry, and government (see Appendix A for participant list). The workshop sought to articulate the potential for non-military drone applications by 2035 and to uncover the critical bottlenecks and challenges that we must overcome to realize that future. This report parses those findings into actionable and coordinated steps with concrete timelines and prioritized recommendations that collectively provide a clear pathway to the envisioned future. Through discussion and consensus building at the workshop, workshop participants articulated a vision for drone development in the next ten years. The recommendations outlined in each topic area below are prioritized, with the first recommendation as the most timely and integral to future developments throughout the field. This report is for policymakers, research program managers, startups and venture capitalists, and researchers.

### *Section 2.1 Alternative Pathways to Impact*

The applications in the previous section reflect the culmination of immediate, coordinated and substantial investments in edge and cloud computing, networking and cybersecurity, and AI foundations for robust, multi-agent, heterogeneous systems. Collectively, the investments described in this report can close the capability gap. However, we also acknowledge the difficulty of tightly coordinated action. Fortunately, with a strong drive to broadly adopt drone technology already in place, alternative outcomes can still be impactful. Consider the following scenarios.

*Scenario A: Edge AI Breakthrough Only*

In this scenario, research and infrastructure investments yield novel systems to manage edge, cloud, and mobile resources across multiple coordinated and seamlessly communicating drones, achieving robust fleet learning, efficient on-drone inference, and formal verification of autonomous behaviors. However, they do not improve trustworthiness, nor do they yield scalable AI ready for production-grade environments. Under this scenario, in 2035, drone applications would have access to democratized aerial sensing for controlled, local deployments with tangible economic value. Example applications include:

- A parent records their child's soccer game from multiple angles with video that is instantly, privately, and securely transmitted and processed for replay from any perspective.
- An agricultural cooperative monitors crop stress across hundreds of acres, reducing water usage by 30% while boosting yields.

*Limitations*: Without mature security frameworks, such applications remain confined to low-stakes environments where breaches carry minimal consequences and single-operator environments where multi-organization collaboration is limited. Without scalable policy frameworks, operations are restricted to private property or specially permitted zones. The technology exists but cannot scale to shared airspace or critical infrastructure protection. For example, advances in edge AI alone would enable fleets of helper drones that communicate with local tractors to assist agricultural workloads. However, without scalable policy and security, the technology is unlikely to reach the 200,000 tractors sold annually in the United States, putting our ambitious goal of scaling to 1 million drones in jeopardy (Odegard, 2025). Edge AI alone creates localized excellence without systemic transformation.

*Scenario B: Security and Trust Breakthrough Only*

Alternatively, imagine security researchers achieve distributed trust mechanisms, formal verification at scale, and robust defense against adversarial drones — but edge AI capabilities and policy frameworks remain immature.

This scenario enables infrastructure-critical operations where drones fly established routes between checkpoints, scanning for emergency situations, identifying unpermitted aerial objects, and delivering select goods. Think utility inspection drones following power line corridors, or security drones patrolling critical facilities on predetermined paths. Human operators monitor thousands of operations hourly through advanced interfaces, ready to intervene when autonomous systems approach verified safety boundaries.

*Limitations*: Without sophisticated edge AI and collaboration via communication, drones cannot adapt dynamically to novel situations, handle complex coordination in contested environments, or learn from fleet experience. Operations succeed through careful engineering of constrained scenarios rather than general-purpose autonomy. Without policy evolution, even secure systems face regulatory barriers limiting where and when they operate. Security alone creates trustworthy but brittle systems that excel in structured environments but cannot handle the full complexity of open-world operations.

*Scenario C: Policy and Testbed Evolution Only*

Finally, consider rapid policy evolution — proactive airspace management frameworks, liability reform, certification pathways for autonomous systems — without corresponding advances in security or edge AI.

This unlocks business opportunities across medium and large-scale enterprises. Clear regulatory frameworks attract investment in drone services: package delivery, infrastructure inspection, agricultural monitoring, emergency response. Coordinated cross-agency initiatives (NSF, USDA, DOT/FAA, FCC, DOD) yield high-impact problems. Liability clarity enables insurance markets. Airspace management systems accommodate increased density.

*Limitations*: Without security foundations, operations remain vulnerable to spoofing, jamming, and adversarial drones. Without edge AI breakthroughs, applications rely on human-intensive operations that limit scalability and increase costs.

These scenarios demonstrate that partial progress can unlock valuable but bounded markets. Only coordinated breakthroughs will achieve the vision of safe, scalable, transformative drone operations in open, shared airspace.

### *Section 2.2 Critical Role of Policy Setting*

The pathways to impact the future of drones present challenges in the American policy and regulatory environment. Even contributions along a single dimension toward computing on the fly will require grappling concurrently with current and evolving policy — e.g., regulations that constrain which research problems matter, which solutions are deployable, and which businesses are viable. Multiple regulatory and standards bodies shape this landscape. The Federal Aviation Administration (FAA) establishes operational rules (Part 107 for small unmanned aircraft systems [UAS], emerging Beyond Visual Line of Sight [BVLOS] frameworks), certification requirements, and airspace management protocols. ASTM International develops consensus standards for the design, performance, and safety of unmanned aircraft systems. The Association for Unmanned Vehicle Systems International (AUVSI) bridges industry, government, and academia to advocate for policies enabling safe UAS integration. Traditional standards bodies like IEEE contribute to underlying infrastructure (wireless communication, cybersecurity), while SAE International and ISO develop drone-specific operational and safety standards.

# 3. Core Challenges and Recommendations

## Scaling to Millions of Drones

The drone ecosystem faces an unprecedented scaling challenge: projections indicate growth from thousands of manually operated aircraft today to millions of autonomous drones by 2035. The U.S. Army seeks massive increases in drone manufacturing, while commercial registrations are forecast to triple by 2035. Such growth will fundamentally transform computational, regulatory, and infrastructure requirements — current centralized systems designed for hundreds of aircraft cannot accommodate millions of simultaneous operations.

### *The Technical Challenge*

This scaling transition exposes four critical bottlenecks that current systems cannot address.

*Spectrum contention*: Dense aerial networks will create severe wireless congestion as millions of mobile nodes compete for limited spectrum, with heterogeneous requirements from low-bandwidth telemetry to high-throughput video streaming. The challenge is compounded by application dependency — different use cases (infrastructure inspection, delivery, surveillance) have fundamentally different data rate, latency, and reliability requirements, yet lack standardized communication protocols. Missing classification schemes for drone types create uncertainty regarding spectrum allocation, priority handling, and interference management as heterogeneous platforms from different operators must interoperate securely and robustly.

*Airspace deconfliction*: Real-time three-dimensional path planning for simultaneous aerial vehicles grows combinatorially complex, requiring distributed coordination mechanisms beyond current centralized air traffic management systems.

*Certification bottlenecks*: Traditional testing approaches become infeasible for millions of autonomous vehicles with continuously updating AI models — testing surfaces expand exponentially as emergent multi-agent behaviors create failure modes unpredictable from single-vehicle analysis.

*Data infrastructure*: Petabyte-scale real-time data from million-drone operations creates fundamental tensions between on-drone generation, distributed edge-cloud processing, and real-time execution requirements. Energy scarcity compounds these challenges — at fleet scale, thermal stress and dynamic voltage/frequency scaling induce latency jitter when control stability margins are thinnest, while synchronized charging contention at mothership docking stations risks cascading mission failures.

### *Catalyzing Recommendations*

- **Create simulation-to-reality pipelines** using digital twin technologies and hybrid simulation approaches combining synthetic and real-world data to validate million-drone behaviors before physical deployment, bridging the sim-to-real gap. [2-4 years]
- **Develop scalable coordination protocols** for distributed consensus, hierarchical fleet coordination, and Byzantine fault tolerance specifically designed for large-scale aerial operations accounting for dynamic network topologies and intermittent connectivity. [3-5 years]
- **Establish spectrum research programs** for cognitive sharing, dynamic access protocols, and interference mitigation in dense aerial networks, coordinating with regulatory bodies (FAA, FCC) to explore dedicated allocations or innovative sharing arrangements. [5-7 years]
- **Build incremental validation frameworks** enabling certification through compositional reasoning, formal verification of critical subsystems, and continuous monitoring rather than exhaustive pre-deployment testing — requiring collaboration between computer science, aerospace engineering, and regulatory communities. [4-6 years]
- **Develop application-specific communication standards** addressing the data rate, latency, and reliability requirements across diverse use cases (delivery, inspection, emergency response, etc.), including classification schemes for drone types that enable priority-based spectrum allocation and interference management. [4-6 years]

## Building Reliable Fleets From Non-Deterministic Agents

Traditional systems' integration practices — unchanged for decades — assume predictable components with well-defined interfaces. These assumptions fail for AI-driven drones where behaviors emerge from learned models rather than explicit programming, creating what workshop participants termed an "integration crisis." When a fleet of delivery drones must coordinate through a thunderstorm, each vehicle's AI makes autonomous decisions — but proving the collective behavior remains safe requires verification techniques that do not yet exist at scale.

### *The Technical Challenge*

Fleets must coordinate non-deterministic AI agents toward common goals, but traditional integration principles expect predictable components with defined interfaces. This mismatch creates four critical problems.

*Non-deterministic decision-making*: Deep learning models may produce different outputs for identical inputs, breaking verification approaches.

*Elusive constraints*: Neural networks may respect constraints during training but violate them in novel operational scenarios.

*Emergent multi-agent behaviors*: Fleet coordination algorithms working in simulation may produce oscillations or deadlocks when deployed with real vehicles, with state spaces growing exponentially with fleet size.

*API opacity*: Integrators increasingly work with black-box AI services, making it impossible to reason about failure modes or verify correct behavior.

### *Catalyzing Recommendations*

- **Establish integration testbeds** (physical + digital) supporting graduated complexity, realistic failure injection, and shared milestones for reproducible research. [2-3 years]
- **Fund integration support platforms** providing runtime monitoring, automated test generation, and safe fallback mechanisms for composing AI agents into safety-critical systems. [3-5 years]
- **Develop Digital Twins 2.0** with bidirectional synchronization, real-time telemetry ingestion, and faster-than-real-time simulation for failure prediction. [4-6 years]
- **Advance formal methods at scale** for neural network verification, probabilistic model checking, and compositional approaches handling multi-million parameter models. [5-8 years]

- **Promote model-based systems engineering for AI** extending frameworks to represent learning processes, uncertainty bounds, and failure mode taxonomies. [3-5 years]
- **Train systems integration workforce** through graduate programs at 20 universities and industry-academia partnerships to address the critical talent gap. [4-7 years]

## AI Intelligence and Assurance

AI-driven intelligence is central to advancing drone capabilities, yet aerial systems impose constraints that fundamentally distinguish them from other AI application domains. Drones operate in safety-critical environments, interact directly with people and infrastructure, and must comply with strict regulatory regimes. Errors are physical, often irreversible, and highly visible. A perception system that misclassifies a stop sign as a yield sign 0.1% of the time may be acceptable for recommendation algorithms but catastrophic for flight control—yet both achieve 99.9% accuracy. Progress in drone AI depends not on abstract autonomy, but on disciplined approaches integrating learning with physical models, certification, human oversight, and verifiable safety guarantees.

### *The Technical Challenge*

Recent advances in generative and learning-based AI have expanded what drones can perceive and reason about, but expose a core tension: AI systems that perform well statistically may fail unpredictably when embedded in physical control loops. Three interconnected problems must be solved together.

*Physics-decoupled learning*: Increasing reliance on data-driven approaches risks decoupling autonomy from physics, control theory, and foundational engineering principles. Current perception pipelines suffer from a 'depth-semantic gap' — successfully recovering geometry while omitting safety-critical physical semantics such as friction, compliance, and load-bearing capacity essential for stable physical contact in close-proximity operations.

*Static certification for dynamic systems*: Traditional certification approaches rely on pre-specified requirements and fixed test suites, creating structural mismatches with learning systems that adapt over time. This forces either stalled deployment or premature operational use without sufficient assurance, as AI systems evolve but certification remains frozen.

*Unverifiable collective intelligence*: Future operations will rely on fleets of heterogeneous drones deployed incrementally, benefiting from shared learning and coordination. However, designing AI systems that support scalable, certifiable collective intelligence without uncontrolled emergent behavior remains an open challenge, particularly when experience gained by one drone affects others' behavior in ways difficult to predict or audit.

### *Catalyzing Recommendations*

- **Invest in hybrid AI architectures** that tightly integrate learning-based perception and decision-making with physics-aware models and control frameworks, treating uncertainty as a first-class system property enabling drones to recognize degraded performance, unknown conditions, and knowledge gaps. [3-5 years]
- **Develop dynamic, evidence-based certification approaches** that enable iterative revision of permissible (use, context) pairs as operational evidence accumulates, shifting toward functional assurance that decomposes autonomy into testable capabilities with assurance levels scaled to potential failure severity. [4-6 years]
- **Integrate runtime monitoring and diagnostics** to detect anomalous behavior, uncertainty, or constraint violations and trigger conservative fallback or human re-engagement, connecting AI outputs to physical margins and operational envelopes rather than just task success rates. [3-5 years]
- **Advance fleet learning architectures** supporting knowledge sharing while preserving independence, auditability, and safety — enabling experience gained by one drone to benefit others without catastrophic interference, with principled mechanisms for transferable collaborative learning and verification of group-level effects. [5-7 years]
- **Co-develop AI assurance standards** across regulators, researchers, and industry, treating AI explicitly as an engineering tool and aligning development, testing, and certification practices to avoid post-hoc rulemaking that stifles innovation. [4-6 years]

## Trust, Security, and Distributed Authentication

Drones are evolving from simple remotely-operated devices into sophisticated cyber-physical platforms hosting AI agents within complex distributed networks. Yet current approaches assume perfect connectivity and rarely account for real-world threats like spoofing, jamming, and interference. As drones integrate into satellite networks, terrestrial communications, and edge computing infrastructures, traditional centralized trust models become inadequate.

### *The Technical Challenge*

Research communities operate in silos, with barriers between drone researchers, networking specialists, AI/edge computing experts, and security professionals. This fragmented approach creates critical vulnerabilities: robotics algorithms assume perfect communication, location spoofing breaks coverage applications, and communication disruptions cascade into system failures. Identification through Remote-ID and emerging standards may still be vulnerable to spoofing by adversaries. Unlike ground vehicles, drones operate with the third dimension of height as a significant consideration, and operate at greater speed than other modalities. The autonomy stack is developed without security considerations, creating fundamental architectural weaknesses. Four problems demand coordinated solutions.

*Distributed trust without centralized authority*: Autonomous agents operating across platforms require continuous attestation and authentication during degraded or blackout communication conditions, yet current solutions like 5G V2X SideLink have poor security implementations.

*Untrusted infrastructure*: Integration with edge computing, satellite networks, and 6G introduces new attack vectors.

*Inter-agent verification*: Autonomous agents must assess trustworthiness when sharing sensor data or delegating subtasks across organizational boundaries.

*Privacy-preserving coordination*: Drones must collaborate (collision avoidance, mission planning) while protecting sensitive operational data, requiring cryptographic protocols that function under the computational and latency constraints of aerial platforms.

### *Catalyzing Recommendations*

- **Establish scalable and performant drone-based entity tags and key infrastructure** for identification and tracking at device level. [2-5 years]
- **Foster interdisciplinary security research and establish shared infrastructure** through forums bringing together robotics, networking, security, and AI communities to support research across adversarial scenarios, communication degradation, and supply chain verification. [2-4 years]
-
- **Develop integrated security frameworks** that assess communication link failures and their cascading effects on autonomy, navigation, and mission execution, incorporating continuous dynamic attestation where drones verify system integrity without centralized authority. [3-5 years]
- **Create privacy-preserving coordination protocols and decentralized trust architectures** enabling secure drone-to-drone collaboration without exposing sensitive operational data, supporting scenarios where drones serve as communication relays and trust anchors. [4-6 years]
- **Implement context-aware security** that adapts authentication requirements and trust models based on operational zones (public, restricted, military airspace) and communication conditions (full connectivity, degraded, blackout). [5-7 years]

## Edge-Cloud Continuum and Real-Time Coordination

Drones operating in close-proximity, safety-critical missions — low-altitude infrastructure inspection, precision agriculture near workers, last-meter delivery in cluttered neighborhoods — serve as the primary point of physical contact where time and space function as hard boundary conditions for stability. A delivery drone navigating between buildings must process sensor data, plan collision-free paths, and execute motor commands within milliseconds; even minor latency jitter can destabilize flight. Yet current autonomy stacks suffer from fundamental mismatches between the rigid timing requirements of physical control and the unpredictable execution characteristics of learning-based AI pipelines. This creates a critical challenge: how to coordinate intelligent decision-making across the drone-edge-cloud continuum while maintaining the real-time guarantees essential for safe physical operation.

### *The Technical Challenge*

Coordinating autonomous fleets across the drone-edge-cloud continuum requires solving problems that monolithic autonomy pipelines cannot address. Four interconnected challenges emerge:

*Real-time execution gaps*: Learning-based perception and planning pipelines rarely provide Worst-Case Execution Time (WCET) guarantees, allowing long-tail latency and jitter to erode control stability precisely when safety margins are thinnest. Opaque execution semantics in common software libraries hinder enforcement of real-time scheduling, bounding of tail latency, and principled compute-energy tradeoffs. Current systems cannot guarantee that critical computations (collision avoidance, motor control) complete within bounded time under worst-case conditions.

*Depth-semantic gaps*: While 2D/3D perception pipelines successfully recover geometry, they often omit safety-critical physical semantics such as friction, compliance, and load-bearing capacity essential for stable physical contact. A drone landing on a rooftop must assess whether the surface can support its weight, not just recognize it as a flat plane. Sensor fusion architectures lack mechanisms to align visual mapping with proprioceptive and inertial sensing to infer functional physical properties and express uncertainty in controller-relevant terms.

*Energy-timing coupling*: Energy scarcity becomes a first-class safety constraint at fleet scale. Thermal stress and dynamic voltage/frequency scaling induce latency jitter when control stability margins are thinnest. Mothership-based charging creates queuing and scheduling interdependencies; naïve battery-threshold policies risk synchronized contention for power and bandwidth, potentially triggering cascading mission failures. Without cross-layer optimization, sustainability and scalability become mutually conflicting objectives.

*Heterogeneous integration brittleness*: Mothership-drone architectures aggregate global situational awareness, perform cross-drone task allocation, and support automated docking, extending mission duration while enforcing system-wide safety constraints. However, multi-vendor deployments introduce brittle heterogeneity through proprietary protocols. The mothership becomes a system-level bottleneck during tightly coupled operations (precision landing on moving platforms under turbulence), while the overall system remains exposed to cyber-physical threats including global navigation satellite system spoofing and jamming. Field-Programmable Gate Arrays (FPGA) offer low-jitter, energy-efficient adaptability but remain underutilized due to tool chain friction and limited integration with mainstream autonomy stacks.

### *Catalyzing Recommendations*

- **Develop real-time-aware autonomy frameworks** exposing explicit timing, energy, and Quality-of-Service interfaces enabling analyzable composition and runtime adaptation, with standardized cross-layer reconfiguration APIs allowing drones and edge nodes to negotiate sensing rates, model fidelity, and control frequencies under enforced safety constraints and predictable timing guarantees. [3-6 years]
- **Invest in physically grounded sensor fusion architectures** aligning 3D visual mapping with proprioceptive and inertial sensing to infer functional physical properties (friction, compliance, load-bearing capacity) and express uncertainty in controller-relevant terms, supported by benchmarks evaluating embodied spatial reasoning under strict on-device computing and energy budgets. [3-5 years]
- **Promote Value-of-Information-driven resource management** replacing static battery thresholds and computation policies with intent-aware approaches allocating energy, effort required for sensing, and computational resources according to mission risk and uncertainty, enabling graceful degradation while preserving control stability and real-time guarantees. [4-6 years]
- **Support FPGA-enabled adaptive computing frameworks** leveraging high-level synthesis and partial reconfiguration to permit hardware functions to be swapped or power-gated within certifiable timing bounds, enabling energy-efficient adaptability without sacrificing real-time predictability. [5-7 years]
- **Develop intent-aware fleet orchestration architectures** treating drones and mothership infrastructure as unified cyber-physical systems with end-to-end timing, safety, and resilience assurances, including safety-critical over-the-air update mechanisms that formally preserve timing contracts during fleet-wide deployment. [5-8 years]
- **Advance privacy-by-design data architectures** emphasizing transmission of task-relevant features and physical semantics instead of raw, identifiable sensor streams, reducing privacy exposure while preserving mission utility and minimizing uplink bandwidth consumption. [5-8 years]

## Next-Generation Drone Networks

In the past, the standard paradigm for drone operations was a strict 1:1 ratio — one human operator piloting one drone. As automation advanced, this evolved into the 1:N model, where a single operator could manage a swarm of multiple drones. Today, the aerospace and robotics industries are moving toward the ultimate paradigm: the **m:N** concept. In an m:N architecture, **m** number of operators (which can include human pilots, ground control stations, cloud-based AI agents, or air traffic control systems) simultaneously interact with, control, and monitor **N** number of drones. It can be foreseen that m will become smaller and N will increase in future applications. The m:N concept represents a paradigm shift from individual (line of sight) drone piloting to scalable, shared, and autonomous fleet management. Its success is fundamentally tethered to the evolution of robust communication networks.

Current standards for the communication of drones, e.g., IEEE 1920.1 (IEEE Standards Association, 2023) are agnostic to the type of network. Communication can be realized either directly — device to device (D2D) — or via some kind of relay or base station, including Public Land Mobile Networks such as 5G and beyond and Satellite Communication. Besides just communication, new technologies — such as 6G with its expected rollout by 2030 — will introduce Integrated Sensing and Communication (ISAC) capabilities, where the same waveform serves both sensing and communication functions. This enables spectrum-efficient operations where drones simultaneously navigate via radar sensing and coordinate with fleet members using the same signal.

### *The Technical Challenges*

In an m:N scenario, networks are not just transmitting simple telemetry. Multiple drones may be simultaneously streaming high-definition video, LiDAR data, ISAC data, and heavy sensor payloads to multiple operators across different geographic locations. This many-to-many, beyond line of sight communication requires massive network throughput. If the network cannot handle the load, bottlenecks will occur, degrading situational awareness for the operators. While sensor data can tolerate slight delays, Command and Control (C2) links cannot. If multiple operators are issuing commands to a fleet — especially in fast-changing or collision-risk scenarios — the network must guarantee ultra-low latency. Furthermore, the synchronization of system states across all *M* control nodes and *N* drones requires constant, instantaneous updates. A dropped packet or high latency could result in conflicting commands, leading to mission failure or a crash.

Drones move at high speeds in three-dimensional space. In an m:N setup, the network topology is constantly shifting. Drones must frequently hand over connections between different cell towers, satellite links, or other drones (via mesh networking) without dropping the C2 link. Maintaining persistent connections between a moving fleet and geographically dispersed operators requires highly advanced routing algorithms and seamless mobility management.

When deploying D2D solutions as well as multi-layered architectures (e.g., High Altitude Platform Stations [HAPS]), the positions of the drones are adaptable (HAPS Alliance, 2026). Different positions and network layouts affect the communication capabilities of the network but also allow for an additional layer of adaptability for the application requirements. Uncertainty (e.g., wind conditions) may affect the path planning for application as well as the network layout. Drones depend on robust and dependable communication architectures that reflect the application specific requirements and platform capabilities on the drones. An m:N fleet may be heterogeneous, utilizing different hardware, operating systems, and communication standards. A major challenge is creating a unified communication architecture (network of networks) and ensuring cross-platform compatibility.

The m:N model vastly expands the attack surface. Because multiple nodes have the authority to issue commands to multiple drones, a malicious actor intercepting just one link could theoretically compromise an entire fleet. The communication network must employ robust, end-to-end encryption and decentralized authentication protocols. Additionally, managing the radio frequency spectrum to prevent unintentional signal interference among the *N* drones and *M* control stations is highly complex in crowded airspaces. Besides, ISAC fundamentally couples sensing and communication in ways that complicate traditional security models. Three interrelated problems emerge.

*Dual-use signal vulnerabilities*: The same waveform used for navigation can be spoofed to create false obstacles or jammed to blind the drone, while communication data embedded in sensing signals may leak operational information to passive observers. A related challenge is achieving centimeter-level local positioning accuracy through joint communication and sensing for both indoor and outdoor environments using passive, privacy-preserving techniques that scale to millions of drones — essential for precise navigation in GPS-denied environments and dense urban operations where traditional positioning fails.

*Performance-security tradeoffs*: Operators resist security measures that degrade functionality, yet encrypting sensing waveforms or adding authentication overhead reduces detection range and increases latency in time-critical flight control.

*Adversarial sensing*: ISAC enables sophisticated counter-drone systems to detect and track unauthorized drones using their own communication signals, but the same techniques allow malicious actors to monitor authorized operations or conduct reconnaissance on critical infrastructure.

### *Catalyzing Recommendations*

- **Create counter-drone detection systems leveraging ISAC** to identify unauthorized drones through passive sensing of their communication signals, enabling protection of critical infrastructure while respecting privacy in public airspace. [3-5 years]
- **Create collaborative, innovation platforms for select use cases** that enable testing, interoperability and system integration to develop secure, robust communication of heterogenous platforms of different customers [3-7 years]
- **Address sensing-communication performance tradeoffs** through spectrum availability and advanced signal processing techniques (adaptive waveforms, cognitive spectrum management) that optimize security, communication throughput, and sensing accuracy dynamically based on operational context. [5-7 years]
- **Utilize drone/flight path/network layout adaptability** in developing mechanisms and approaches for future applications. [5-8 years]
- **Develop security protocols for ISAC implementations** that protect sensing waveforms from spoofing and jamming while maintaining detection performance and enabling centimeter-level positioning accuracy in both indoor and outdoor environments through passive, privacy-preserving techniques. [4-6 years]

## AI Autonomy and Agentic Systems

The emergence of LLM-enabled agents on UAVs creates new challenges for autonomous decision-making in dynamic environments. Future architectures must decouple agents from specific platforms, allowing them to migrate between drones, edge servers, and cloud resources as mission requirements change — a wildfire-monitoring agent might start on a ground station, transfer to a drone for aerial survey, then migrate to edge infrastructure for data processing. However, this flexibility introduces new attack vectors and resource management challenges that traditional fixed-deployment models do not address.

### *The Technical Challenge*

Platform-agnostic agents require fundamental rethinking of execution guarantees and resource management across heterogeneous computing environments. Two critical problems emerge.

*Secure agent migration*: Transferring AI agents between platforms (drone processors, edge servers, cloud) while maintaining code integrity, protecting model weights and training data, and preventing adversaries from injecting malicious behavior during transitions. Current security models assume agents remain on trusted hardware throughout their lifecycle.

*Cross-platform execution guarantees*: Ensuring agents produce consistent, verifiable behavior regardless of underlying hardware, accounting for vastly different computational capabilities (edge TPUs vs. drone microcontrollers), energy constraints (battery-powered vs. grid-connected), and real-time requirements (flight control vs. analytics) across the drone-edge-cloud continuum.

### *Catalyzing Recommendations*

- **Develop platform-agnostic agent architectures** with standardized interfaces enabling secure execution across drone processors, edge infrastructure, and cloud resources while maintaining performance guarantees and security properties. [3-5 years]
- **Create secure migration protocols** for transferring agents between platforms, including cryptographic verification of code integrity, encrypted state transfer, and protection against injection attacks during transitions between trust domains. [4-6 years]
- **Design adaptive resource management systems** that dynamically allocate computational tasks across the drone-edge-cloud continuum based on real-time constraints (latency, energy, bandwidth), mission priorities, and security posture. [4-6 years]

## Human-AI Partnership and Scalable Oversight

As autonomy scales, the role of humans shifts from direct control to supervision, exception handling, and policy enforcement. The transition requires new interface paradigms supporting "human on the loop" rather than "human in the loop" control — a single operator might supervise fifty delivery drones, intervening only when systems encounter novel situations beyond verified competencies. However, poorly designed AI systems risk either overwhelming operators with alerts or removing them from meaningful engagement. For drones operating near people, infrastructure, and shared airspace, this challenge is amplified by safety criticality and public visibility.

### *The Technical Challenge*

Effective human-AI teaming for large-scale drone operations faces three critical challenges.

*Semantic gaps in communication and understanding*: Humans and AI systems often reason about tasks using incompatible representations. An operator thinks "deliver to the north entrance if weather is clear, otherwise wait," but current systems require this translated into low-level parameters that obscure intent and make supervision brittle when conditions change. The need to recognize gaps in knowledge is essential. Natural language interfaces offer promise but introduce security vulnerabilities requiring careful protocol design.

*Supervision scalability*: As fleets grow from dozens to thousands of vehicles, traditional monitoring (one screen per drone, manual intervention for exceptions) becomes cognitively infeasible. Systems must surface the right information without overwhelming operators, yet current interfaces expose either too much detail (drowning operators in telemetry) or too little (hiding critical uncertainty until failures cascade).

*Re-engagement dynamics*: Graded autonomy must allow humans to re-enter control loops during emergencies without destabilizing system behavior, but transitions between autonomy levels create dangerous transient states where neither human nor AI maintains full situational awareness.

#### *Catalyzing Recommendations*

- **Develop shared representations and communication protocols** allowing humans and AI systems to reason about tasks, constraints, and uncertainty using compatible abstractions, including natural language interfaces with security controls against prompt injection and misinterpretation. [3-5 years]
- **Design autonomy systems that augment human capability** enabling small operator teams to supervise large drone fleets safely through intelligent information filtering, uncertainty quantification, and proactive exception flagging. [4-6 years]
- **Create interfaces that expose evidence, uncertainty, and rationale** rather than just system outputs, helping operators understand AI decision-making and identify when systems approach boundaries of verified competence. [3-5 years]

### Data, Training, and Validation Infrastructure

Drone AI depends on data that is difficult to collect at scale due to cost, risk, and regulatory constraints. Flight testing is expensive, dangerous, and restricted to approved airspace. Real-world failure modes (sensor degradation, adversarial attacks, extreme weather) cannot be safely induced during data collection. Synthetic and virtual environments offer promise for training and testing, but must be used carefully to avoid reinforcing biases or masking failure modes that emerge only in physical deployment. The challenge is creating validation infrastructure that reflects the interdependence of vehicle dynamics, environmental conditions, and mission constraints without requiring prohibitively expensive physical testing.

#### *The Technical Challenge*

Generating sufficient high-quality training data and validation evidence for safety-critical drone AI requires solving the simulation-to-reality gap. Three problems constrain current approaches.

*Sim-to-real transfer failures*: AI models trained purely in simulation often fail catastrophically when deployed on physical drones due to unmodeled dynamics (wind turbulence, motor latency, sensor noise) and subtle environmental factors (lighting conditions, material properties, GPS multipath). Bridging this gap requires not just higher-fidelity simulation, but understanding which discrepancies matter for safety versus performance.

*Opportunistic datasets*: Current drone datasets are collected ad-hoc from available platforms and environments, creating coverage gaps that leave AI systems vulnerable to novel conditions. Systems must be designed to collect useful, auditable data by design during operational deployment, but doing so raises privacy, security, and regulatory concerns.

*Static validation*: Traditional testing assumes fixed environments and requirements, but drones operate in dynamic conditions (weather, traffic, infrastructure changes) that evolve faster than validation cycles. Long-horizon experimentation under evolving conditions is essential but requires infrastructure that does not yet exist at scale.

#### *Catalyzing Recommendations*

- **Invest in blended physical-virtual environments** supporting closed-loop training, testing, and validation under realistic constraints, using hybrid approaches that combine high-fidelity simulation with physical-in-the-loop components (actual sensors, flight controllers, communication systems). Include standardized datasets of real-world failure modes (motor vibration signatures, spoofing/jamming traces, adverse-weather edge cases) and resilience-oriented evaluation metrics prioritizing fault containment, recovery latency, and stability under stress rather than best-case accuracy alone. [3-5 years]
- **Develop data collection frameworks** enabling drone systems to generate useful, auditable training data during operational deployment through privacy-preserving logging, federated learning approaches, and regulatory sandboxes permitting expanded data collection for safety research. [4-6 years]
- **Establish long-horizon experimentation infrastructure** enabling repeated evaluation under evolving environmental and operational conditions, treating the environment itself as part of the autonomous system rather than an external static assumption. [5-7 years]

### Standards, Certification, and Regulation

Current security standards are largely process-based rather than outcome-focused, creating compliance burdens without necessarily improving security. The democratization of drone development — including custom-built drones — challenges traditional certification models designed for manufactured products. Supply chain concerns, particularly regarding drones manufactured in countries offering superior functionality at lower costs but uncertain security guarantees, raise sovereignty and security questions. The lack of standardized regulations for UAVs, comparable to automotive cybersecurity standards, creates gaps in security assurance as drone operations scale.

### *The Technical Challenge*

Existing certification frameworks cannot accommodate the pace of software updates in AI-driven autonomous systems. Three problems demand new approaches.

*Process vs. outcome standards*: Current regulations mandate specific development processes rather than measurable security outcomes, creating expensive compliance overhead while allowing functionally insecure systems to pass certification.

*Custom and open-source systems*: Hobbyist and research drones built from commodity components fall outside traditional certification pathways, yet increasingly operate in shared airspace with certified commercial systems.

*Continuous evolution*: AI models update frequently based on new training data or operational experience, but each update theoretically requires recertification under current frameworks, creating an impossible bottleneck for learning systems.

### *Catalyzing Recommendations*

- **Develop scalable self-certification frameworks** for custom drone builders and research platforms, establishing tiered certification levels based on operational risk (private property vs. shared urban airspace) with appropriate testing requirements. [2-4 years]
- **Adapt automotive cybersecurity regulations** (UN R155, ISO/SAE 21434) to drone applications, creating outcome-based security standards that verify measurable security properties rather than mandating specific development processes. [3-5 years]
- **Create continuous certification mechanisms** enabling ongoing validation of AI-driven systems through runtime monitoring, periodic audits, and formal verification of update processes rather than full recertification for each software change. [4-6 years]
- **Establish domestic drone manufacturing incentives** addressing sovereignty concerns through trusted supply chains, secure-by-design hardware architectures, and verification of foreign-manufactured components. [5-7 years]

## Critical Infrastructure Protection

Drones represent a new attack vector against critical infrastructure, with potential for sophisticated attacks using autonomous systems as delivery mechanisms for explosives, surveillance equipment, or cyber payloads. The dual-use nature of drone technology shows both defensive and offensive capabilities at scales that evolve faster than defensive countermeasures. A coordinated attack on a power substation, water treatment facility, or data center could cause cascading failures affecting millions, yet current detection and mitigation strategies are insufficient for addressing the scale and sophistication of potential drone-based threats.

### *The Technical Challenge*

Protecting critical infrastructure from adversarial drones requires solving problems distinct from general drone security. Three challenges emerge.

*Detection in cluttered environments*: Identifying malicious drones near airports, power plants, or stadiums requires distinguishing threats from authorized operations (delivery drones, news helicopters, hobbyists) in real-time, often in environments with significant electromagnetic interference and physical obstructions that degrade sensor performance.

*Attribution and forensics*: After an incident, investigators must determine drone origin, operator identity, flight path, and payload, but current systems lack standardized logging, tamper-evident black boxes, or chain-of-custody protocols for drone-collected evidence.

*Proportional response*: Countermeasures must neutralize threats without collateral damage (jamming disrupts legitimate communications, kinetic intercepts risk debris falling on populated areas) while operating within legal frameworks that vary by jurisdiction and protected facility type.

### *Catalyzing Recommendations*

- **Establish operational protocols defining authorized drone operations** near critical infrastructure, including geofencing requirements, identification broadcasting, and coordination with facility security, backed by clear legal frameworks for enforcement. [2-5 years]
- **Develop proactive threat detection systems** integrating multi-modal sensors (radar, RF, acoustic, optical) with AI-based classification to identify malicious drone activities in cluttered environments, providing early warning for critical infrastructure operators. [3-5 years]
- **Create forensic capabilities for drone incident investigation** including standardized flight data recorders, tamper-evident logging, and evidence collection protocols enabling post-incident attribution and prosecution. [2-4 years]
- **Develop proportional counter-drone technologies** that can neutralize threats in contested environments through targeted jamming, spoofing, or capture mechanisms that minimize collateral impact on legitimate operations and public safety. [4-6 years]

## Workforce Development and Education

Drones are evolving from hobbyist tools into critical infrastructure components spanning logistics, inspection, emergency response, agriculture, and defense. Yet workforce preparation has not kept pace. Between December 2024 and December 2025, approximately 35,000 job postings mentioned drones (spanning film, photography, surveying, nondestructive testing, construction, agriculture), with 5,000 specifically requiring drone computing skills in robotics, software development, and systems integration (Lightcast.io, 2026). However, no coherent educational pathway produces graduates with the interdisciplinary expertise these roles demand — combining AI/autonomy, embedded systems, cybersecurity, aerospace engineering, and regulatory knowledge. Rapid AI advances compound the challenge; over-specialization and "plug-and-play" AI use risks eroding foundational skills in physics and control theory, while insufficient AI literacy among operators increases operational risk.

### *The Technical Challenge*

Drone computing sits at disciplinary intersections, creating institutional friction and educational gaps. Three problems emerge.

*Unclear departmental ownership*: Is drone computing robotics, embedded systems, aerospace, or computer science? No clear answer exists, resulting in fragmented curricula where students piece together knowledge across departments without coherent integration. Universities have not systematically integrated drone skills as foundational competencies despite growing use by civil, environmental, and agricultural engineers.

*Foundational skills erosion*: AI tools enable rapid prototyping but may allow engineers to deploy systems without understanding underlying physics, control theory, or failure modes. FAA Part 107 Remote Pilot Certification provides operational literacy but not the systems engineering, cybersecurity, or AI verification expertise needed for autonomous fleet operations.

*Security and certification deficit*: Limited training exists in UAV-specific cybersecurity with no scalable pathways for training engineers to validate, debug, and analyze failures in autonomous fleets.

### *Catalyzing Recommendations*

- **Reinforce fundamentals-based education** emphasizing physical reasoning, control theory, and systems design alongside ethics, AI literacy, training engineers to validate, debug, and conduct failure analysis on learning-based systems. [Immediate, ongoing]
- **Develop curricula on the ethical use of drones and computing on the fly** emphasizing privacy, public-good tradeoffs, and established frameworks for ethical reasoning. [Immediate, ongoing]

- **Integrate drone computing into engineering curricula** treating drone skills as foundational competencies across disciplines, embedding FAA licensing preparation and regulatory awareness into degree requirements. [3-5 years]
- **Develop interdisciplinary graduate programs** in autonomous systems combining AI/ML, aerospace engineering, cybersecurity, and control theory. [4-7 years]
- **Create shared educational infrastructure** including accessible testbeds, standardized datasets, and simulation environments enabling hands-on experience without requiring expensive flight testing. [2-4 years]

# 4. Roadmap

The technical challenges detailed in Section 3 require timely, interconnected action. Progress across all dimensions creates multiplicative value that cannot be achieved through isolated advances. In this section, we outline key milestones that policy makers, researchers, and industry professionals can reference to benchmark progress in their area while also tracking progress collectively. Section 4.1 lists the milestones by area. Section 4.2 presents concrete recommendations for stakeholders. These milestones and recommendations are the result of technical discussions emanating from the workshop and reflect the consensus of subject matter experts.

## 4.1 Critical 2028 Milestones

The 2035 vision articulated in Section 2 depends on achieving specific technical milestones by 2028. These milestones represent the minimum viable progress needed to unlock the three pathways to impact — edge AI breakthroughs, security and trust advances, and policy evolution — and position the United States for coordinated progress across all dimensions. Without meeting these 2028 targets, the multiplicative benefits described in Section 2 become unattainable.

### *Enabling Edge AI Breakthroughs by 2035*

**Outcome Milestones:**

- Demonstrate autonomous fleet operations (10-50 drones) conducting collaborative tasks (agricultural monitoring, infrastructure inspection) with real-time coordination and graceful degradation under sensor failures, achieving 99% mission completion without human intervention in controlled environments.
- Validate hybrid AI architectures integrating physics-aware models with learning-based perception on representative drone platforms, showing quantifiable improvements in safety margins (30%+ reduction in constraint violations) compared to pure end-to-end learning baselines.

**Capability Milestones:**

- Runtime verification tools capable of monitoring neural networks with up to 100K parameters for constraint violations, providing alerts within 10ms latency suitable for in-flight safety interventions.
- Fleet learning frameworks enabling knowledge transfer across 5-10 heterogeneous drone platforms without catastrophic interference, demonstrated through shared operational datasets from multiple institutions.

**Infrastructure Milestones:**

- National testbed consortium operational at minimum of 10 sites including universities, government labs, and industry partners with standardized APIs, shared datasets including 1,000+ hours of failure mode recordings (sensor degradation, adversarial attacks, extreme weather), and digital twin capabilities supporting 1,000+ drone simulations.
- Blended physical-virtual training environments deployed at 20 universities, enabling students to validate autonomous behaviors in high-fidelity simulation before physical flight testing.

### *Enabling Security and Trust Breakthroughs by 2035*

**Outcome Milestones:**

- Demonstrate secure drone operations maintaining mission integrity under realistic cyber-physical attacks (GPS spoofing, jamming, malicious communication injection) with automatic threat detection and safe fallback responses within 100ms.
- Design, implement, and validate distributed trust mechanisms enabling 20+ drones from multiple organizations to coordinate securely without centralized authority, operating through degraded communication conditions (50% packet loss, 5-second blackouts), realizing ongoing efforts on the Drone Remote Identification Protocol.

**Capability Milestones:**

- Continuous dynamic attestation protocols implemented and tested on commercial drone hardware, verifying system integrity every 100ms with <5% computational overhead.
- Privacy-preserving coordination protocols enabling collaborative flight planning and collision avoidance while protecting sensitive operational data, demonstrated through cryptographic verification under drone computational constraints (sub-watt power budgets).

**Infrastructure Milestones:**

- Community-driven security datasets capturing 500+ adversarial scenarios (spoofing signatures, jamming patterns, supply chain compromises) available to researchers through standardized testbed access.
- Integration security frameworks deployed at 10+ institutions, providing automated analysis of communication link failures and downstream effects on autonomy, navigation, and mission execution.

### *Enabling Policy Evolution by 2035*

**Outcome Milestones:**

- Evidence-based certification frameworks piloted with FAA for autonomous systems, enabling iterative revision of operational permissions based on accumulated flight data from 1,000+ operational hours across diverse scenarios.
- Regulatory sandboxes operational in 3+ states, permitting expanded autonomous operations (beyond visual line of sight, over people, urban environments) under monitored conditions with clear liability frameworks.

**Capability Milestones:**

- Outcome-based security standards adapted from automotive cybersecurity regulations (UN R155) to drone applications, with measurable verification criteria deployed in at least one commercial certification pathway.
- Self-certification frameworks for custom/research drones, establishing tiered certification levels based on operational risk with appropriate testing requirements and automated compliance checking.

**Infrastructure Milestones:**

- Cross-agency coordination mechanisms established between NSF, USDA, DOT/FAA, FCC, and DOD with coordinated research solicitations targeting high-impact applications (wildfire detection, infrastructure monitoring, rural healthcare delivery).
- Workforce development programs launched at 20 universities integrating FAA licensing, systems integration training, and interdisciplinary drone computing curricula, graduating first cohorts (200+ students) with combined AI-aerospace-security expertise.

### *Cross-Cutting 2028 Milestones*

**Integration Demonstration:**

- First integrated demonstration of 1,000-drone coordination combining edge AI (autonomous decision-making), security (distributed trust), and policy (operating under regulatory sandbox) in a realistic mission scenario (disaster response simulation, infrastructure monitoring campaign).

**Standards and Interoperability:**

- Open standards for drone-edge-cloud coordination adopted by 5+ major vendors, enabling interoperable integration platforms that reduce development costs by demonstrable 20%+ margins.

**Real-Time Guarantees:**

- Real-time-aware autonomy frameworks providing WCET guarantees for critical control loops deployed on 3+ commercial drone platforms, enabling formal verification of timing contracts under worst-case sensor and communication conditions.

**Meeting these 2028 milestones positions the United States to:**

- Unlock the $5-15B markets identified in Section 2's individual breakthrough scenarios
- Progress toward the $128B global market requiring coordinated advances across all dimensions
- Maintain technological leadership as international competitors pursue autonomous systems markets
- Address national priorities including the $150B annual weather disaster burden, labor shortages in critical sectors, and national security requirements for autonomous systems at scale

**Failure to meet these milestones by 2028 risks:**

- Ceding technological leadership to international competitors investing aggressively in autonomous systems
- Widening gap between laboratory innovation and deployment-ready trustworthy autonomy
- Missed economic opportunities as regulatory uncertainty and technical limitations constrain commercialization
- Inability to address urgent national challenges (infrastructure deterioration, natural disasters, rural healthcare access) where autonomous drone solutions could provide transformative impact

## 4.2 Stakeholder Recommendations

**For Research Funders:**

Near-peer competitors are out-investing the U.S. in autonomous systems' integration, threatening American leadership in a foundational technology layer. Cross-agency coordination is essential to accelerate autonomous systems' integration and maintain American leadership in a foundational technology layer. NSF should lead foundational research while mission agencies (USDA, DOT/FAA, FCC, DOD) ensure applications address real-world constraints and national priorities. Establish a national testbed consortium enabling reproducible research across diverse platforms and environments, serving 20 universities and 100+ industry partners. Fund integration platforms that compose AI agents into safety-critical systems.

**For Researchers:**

Break disciplinary silos to address the systems integration talent gap. Fewer than a dozen graduate programs currently produce engineers with combined AI-aerospace-security expertise, yet the DOD Replicator Initiative alone will require thousands of such specialists. Autonomous drone research demands collaboration between traditionally separated communities in robotics, networking, AI, security, and human-computer interaction. Engage with real deployments to understand practical constraints — particularly the labor shortage in aviation and infrastructure inspection that creates immediate demand for autonomous solutions. Share datasets and frameworks to accelerate collective progress and avoid duplicative effort across institutions.

**For Industry:**

Open standards enabling interoperability could reduce integration costs by 40-60% based on precedents from wireless and automotive sectors, accelerating time-to-market and enabling smaller companies to participate. U.S. industry participation is essential to maintain technological leadership as international competitors aggressively pursue autonomous systems markets — particularly in Asia and Europe where regulatory experimentation is advancing rapidly. Participate in testbed consortia to validate approaches on real hardware and influence standards development. Share integration experiences and failure mode learnings that can inform research directions and prevent others from repeating costly mistakes. Invest in workforce development for the interdisciplinary systems engineers this future requires, partnering with universities to shape curricula around industry needs.

Critically, domestic drone production must scale. Within the United States, few companies produce programmable consumer drones capable of adopting the computing-on-the-fly technologies discussed throughout this report. Working with policymakers, industry must commit to end-to-end pipelines for development, deployment, and low-cost operation of drones. Expanding domestic manufacturing capacity strengthens supply chain resilience, addresses sovereignty concerns, and enables the rapid iteration and learning essential for competitive advantage in this emerging domain.

**For Policymakers:**

Adopt proactive rather than reactive frameworks across the three pillars: Actor (operator licensing and training), System (vehicle certification and maintenance), and Infrastructure (traffic management, spectrum allocation, airspace design). Safe experimentation frameworks are essential to enable solutions addressing the $150B annual weather disaster burden while maintaining public safety. Regulatory sandboxes can accelerate innovation while managing risk. Coordinate internationally to ensure U.S. leadership in this strategic domain — policy harmonization with allies prevents market fragmentation while maintaining security advantages over adversaries. Finally, as stated above, policymakers must prioritize regulatory environments conducive to drone production and operation at scale. Initial stakeholder sessions detailing the current bottlenecks and barriers could inform critical decisions in the coming years.

## 4.3 Timelines

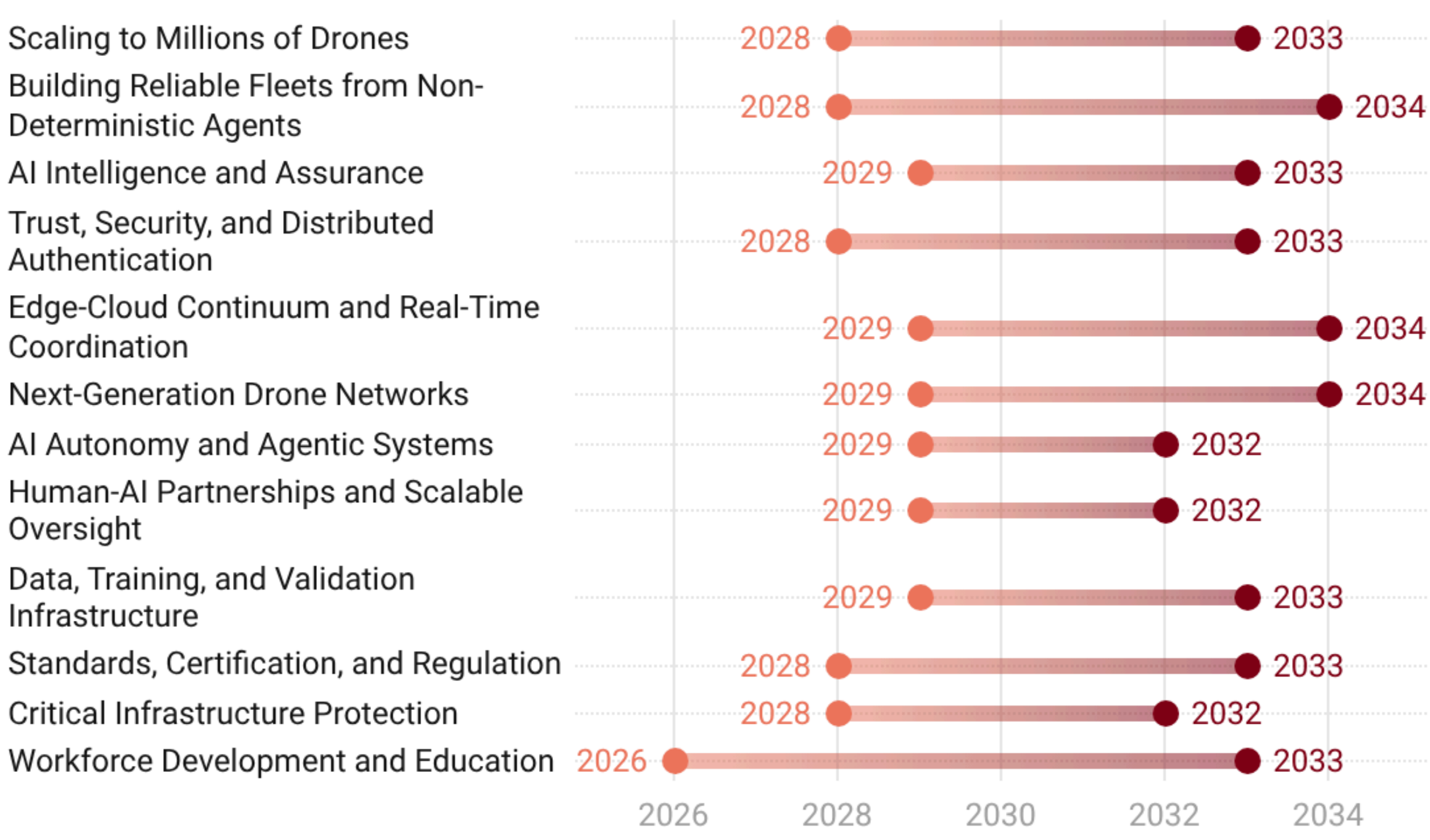

### Scaling to Millions of Drones

Create simulation-to-reality pipelines
Develop scalable coordination protocols
Build incremental validation frameworks
Develop application-specific communication standards
Establish spectrum research programs

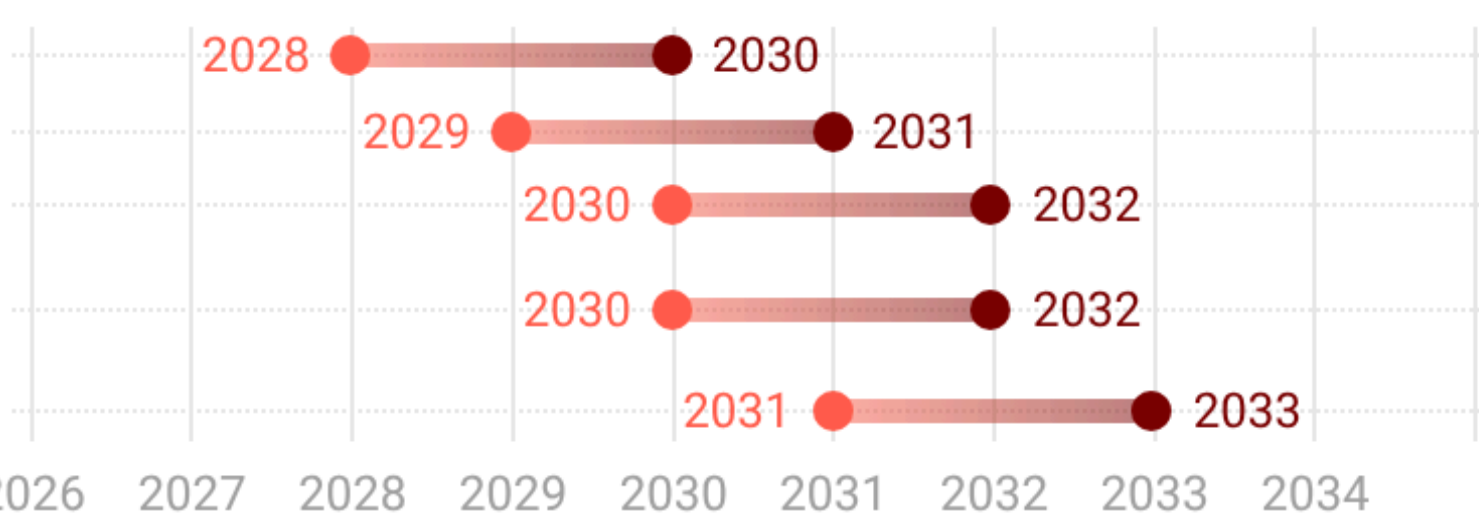

### Building Reliable Fleets from Non-Deterministic Agents

Establish integration testbeds
Fund integration support platforms
Promote model-based systems engineering for AI
Develop Digital Twins 2.0
Train systems integration workforce
Advance formal methods at scale

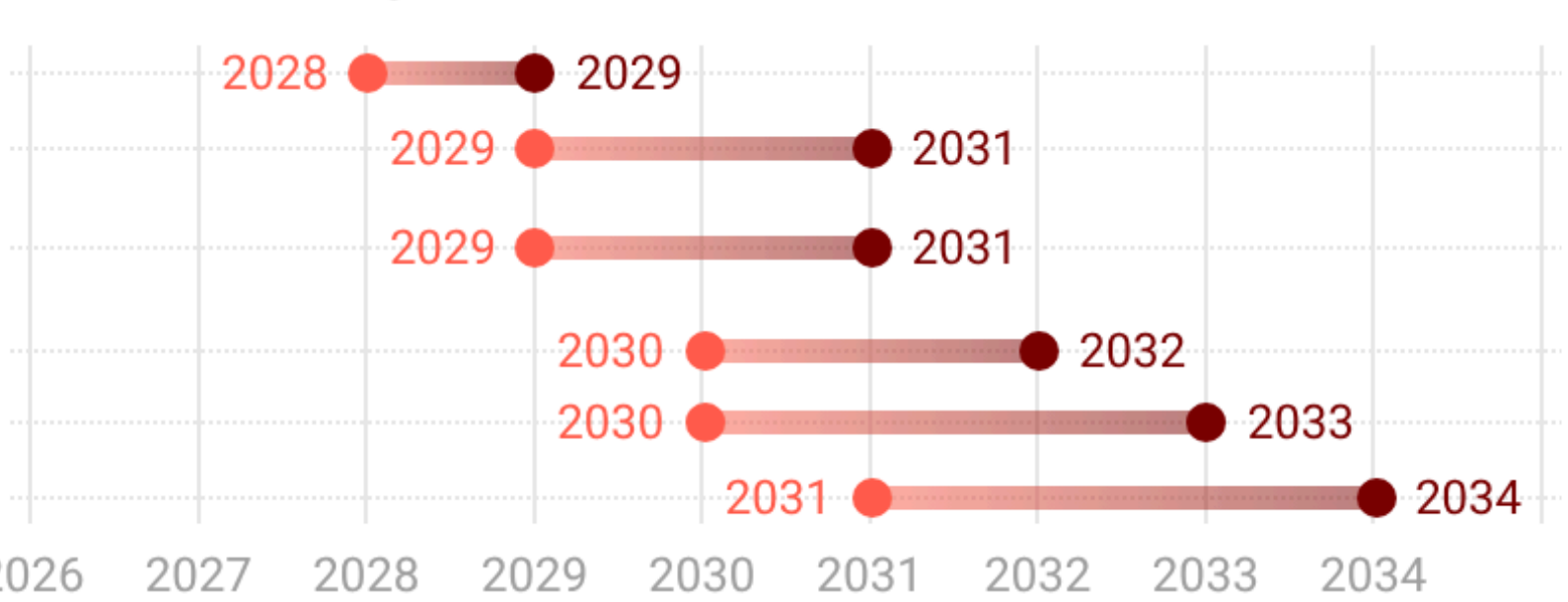

### AI Intelligence and Assurance

Invest in hybrid AI architectures
Integrate runtime monitoring and diagnostics
Co-develop AI assurance standards
Develop dynamic, evidence-based certification approaches
Advance fleet learning architectures

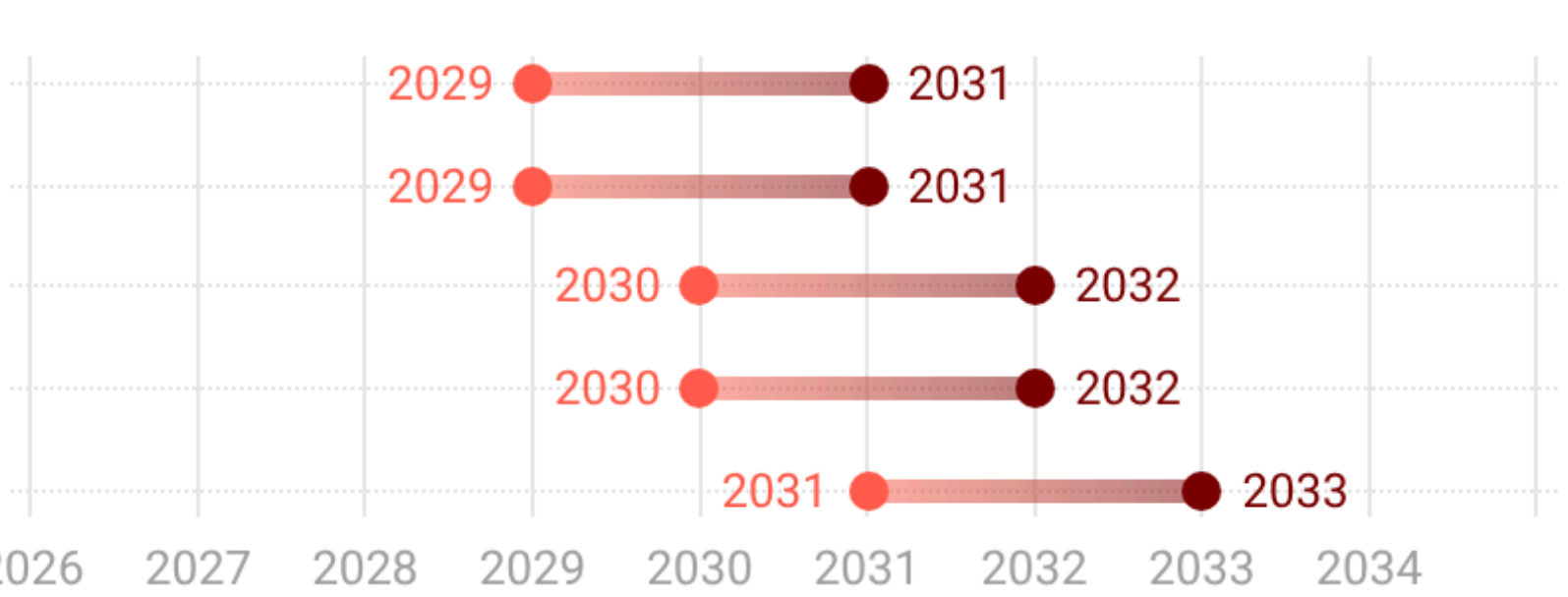

### Trust, Security, and Distributed Authentication

Establish scalable and performant drone-based entity tags and key infrastructure
Foster interdisciplinary security research and establish shared infrastructure
Develop integrated security frameworks
Create privacy-preserving coordination protocols and decentralized trust architectures
Implement context-aware security

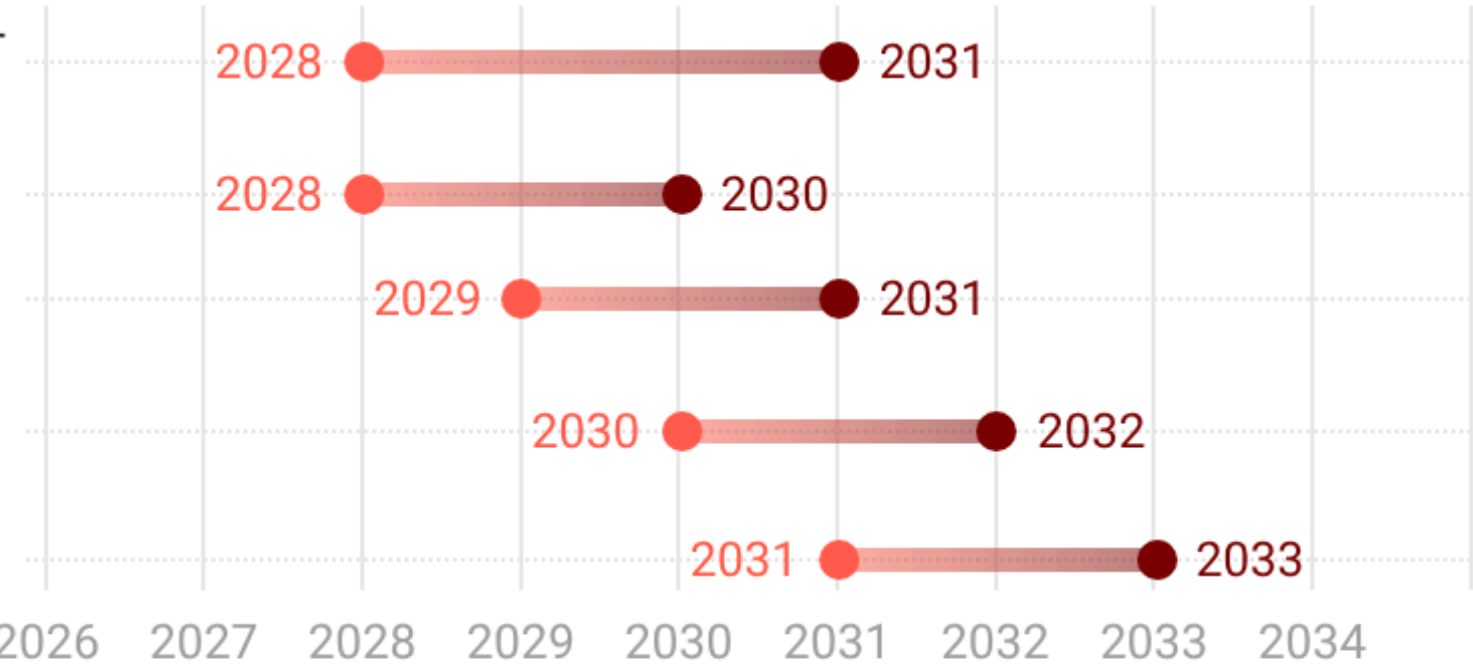

## Edge-Cloud Continuum and Real-Time Coordination

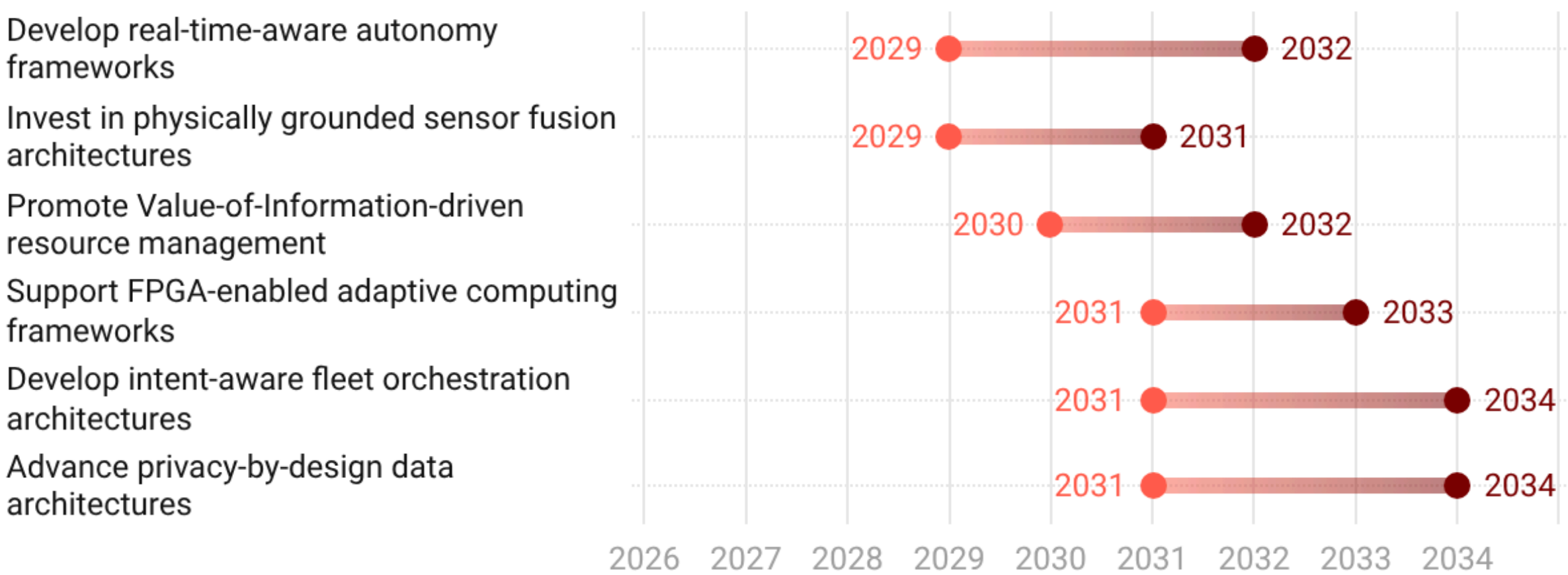

## Next-Generation Drone Networks

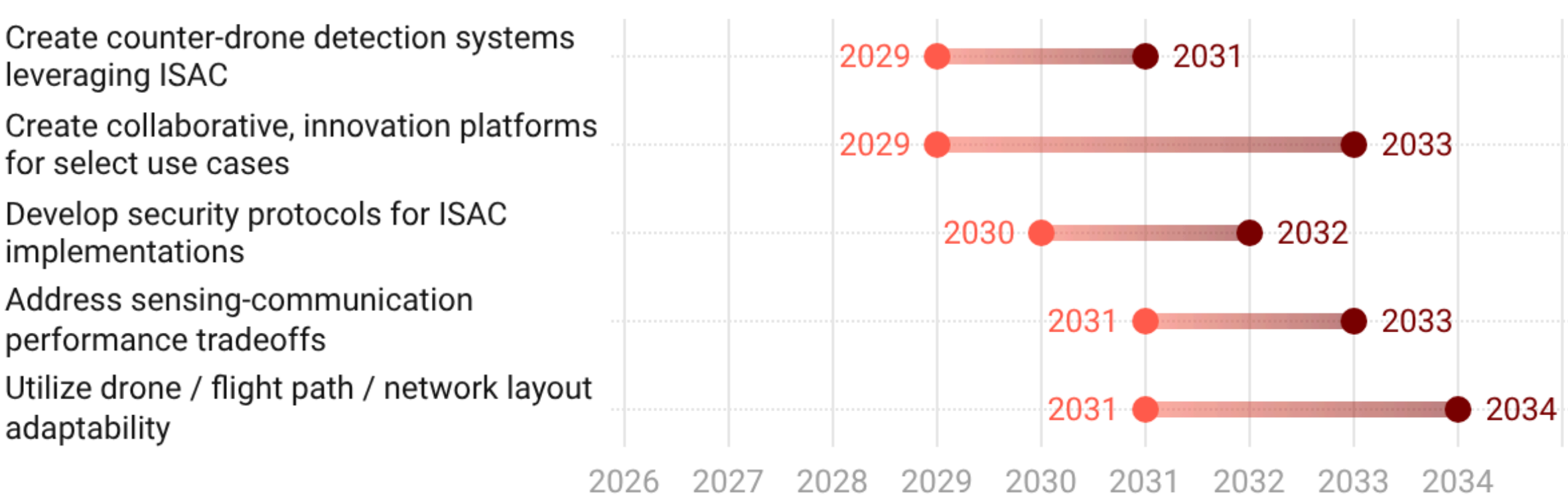

## AI Autonomy and Agentic Systems

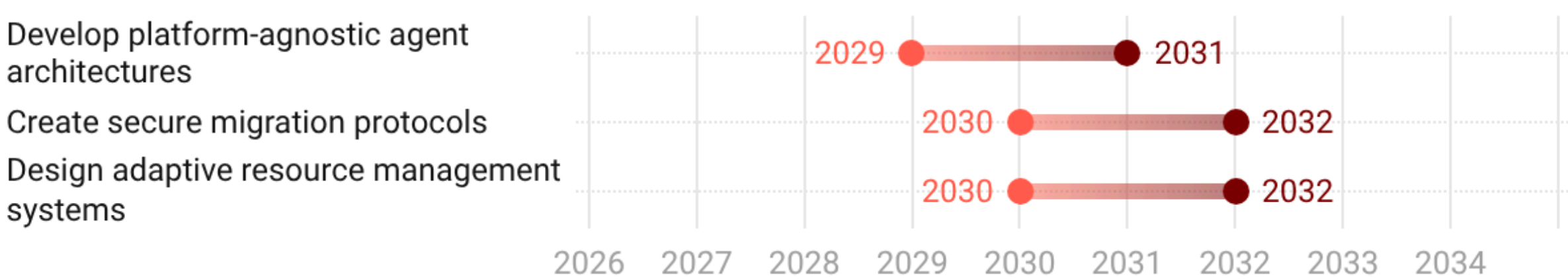

## Human-AI Partnerships and Scalable Oversight

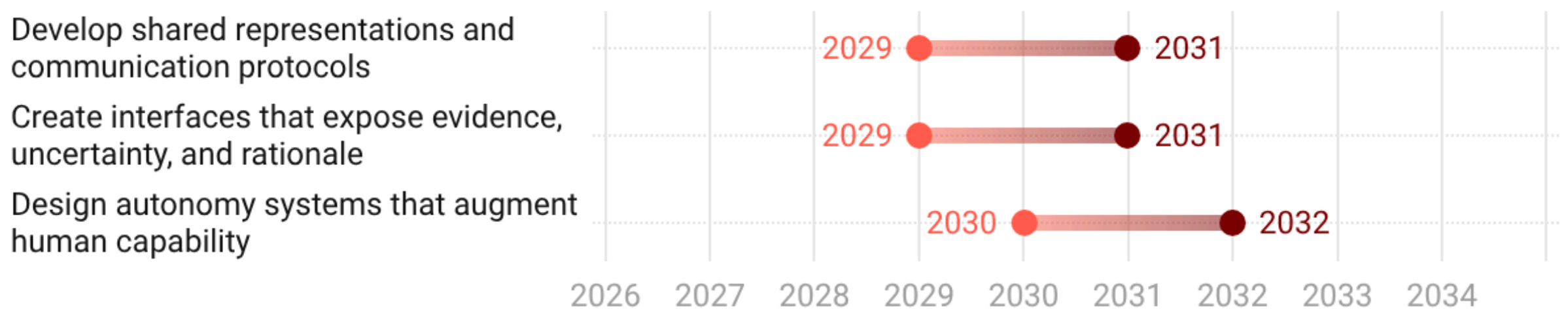

### Data, Training, and Validation Infrastructure

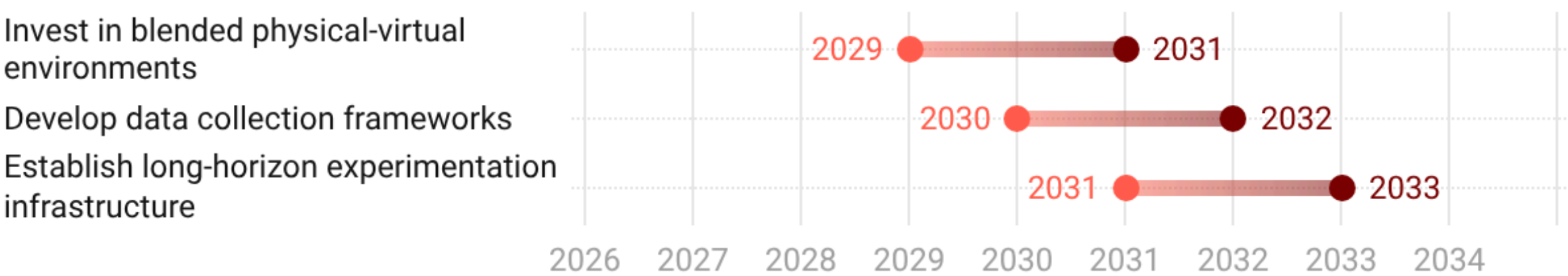

### Standards, Certification, and Regulation

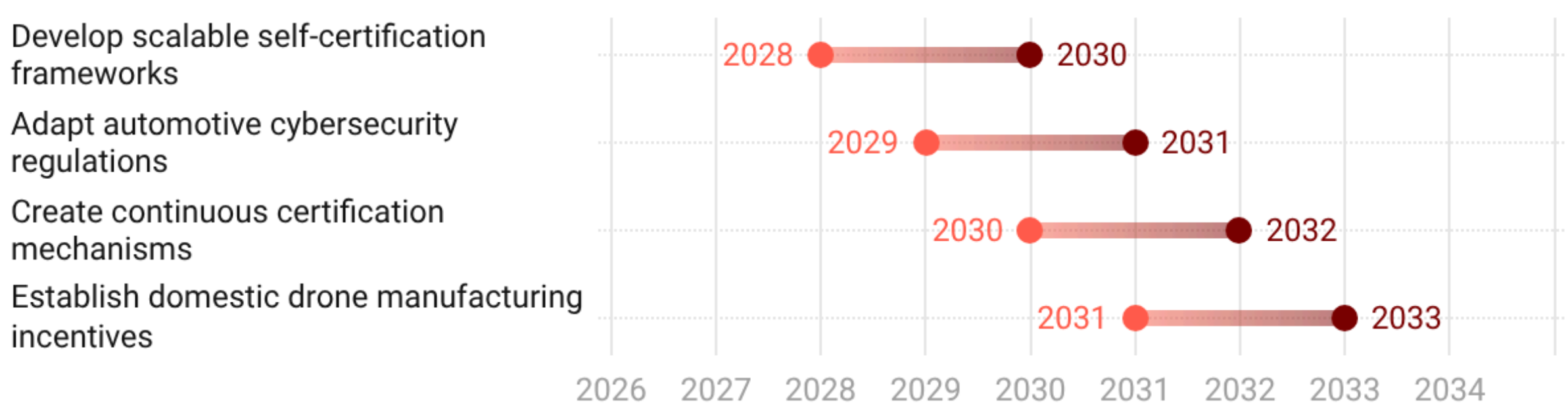

### Critical Infrastructure Protection

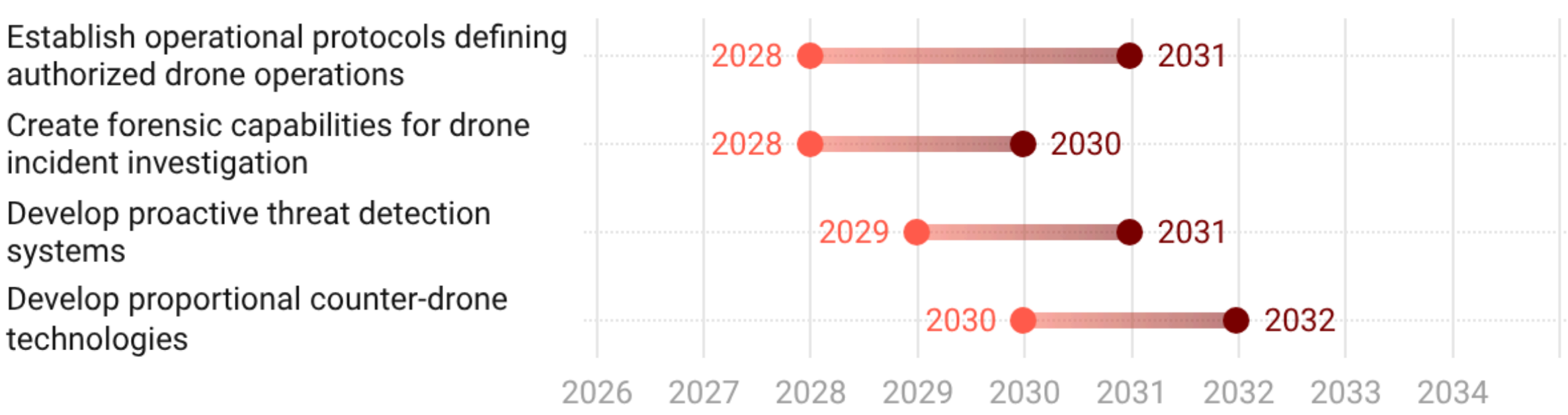

Created with Datawrapper

### Workforce Development and Education

Reinforce fundamentals-based education
Develop curricula on the ethical use of drones and computing on the fly
Create shared educational infrastructure
Integrate drone computing into engineering curricula
Develop interdisciplinary graduate programs

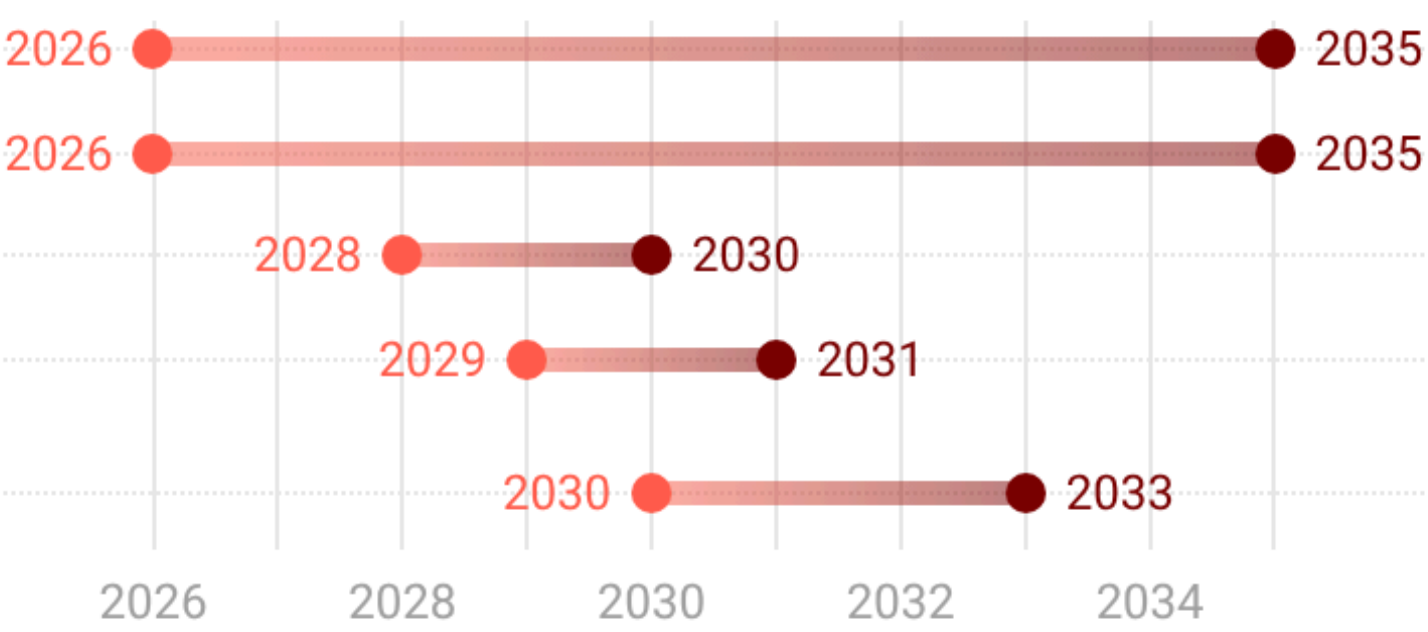

# 5. Glossary

**6G (Sixth Generation Wireless)** The next generation of cellular network technology expected by 2030, offering sub-millisecond latency, micro-second synchronization, and integrated sensing and communication (ISAC) capabilities that will enable drones to simultaneously navigate and communicate using the same waveform.

**AI-Powered Drones** Unmanned aircraft controlled by software that incorporates machine learning models, neural networks, computer vision, and other AI technologies. The software may run on the drone, on nearby edge computing devices, or in the cloud.

**API (Application Programming Interface)** A standardized interface allowing different software systems to communicate. In drone contexts, API opacity refers to the challenge of integrating black-box AI services where internal behavior cannot be inspected or verified, analogous to the "Mechanical Turk problem."

**Attestation** The process of verifying the integrity and trustworthiness of a system's software and hardware components. Continuous dynamic attestation enables drones to repeatedly verify system integrity during flight without relying on centralized authorities.

**Autonomous Systems** Systems capable of performing tasks and making decisions without continuous human control. In drone contexts, autonomy ranges from basic stabilization to full mission planning and execution with fleet coordination.

**Beyond Visual Line of Sight (BVLOS)** Drone operations where the aircraft flies beyond the pilot's direct visual observation, requiring enhanced autonomy, communication systems, and regulatory approval. Critical for scaling commercial operations.

**Byzantine Fault Tolerance** The ability of distributed systems to continue operating correctly even when some components fail or behave maliciously. Essential for drone fleets operating across organizational boundaries with potentially compromised nodes.

**Certification** The formal process of verifying that a system meets safety and performance requirements before operational deployment. Traditional static certification is being challenged by learning-based systems that evolve over time, requiring dynamic, evidence-based approaches.

**Compositional Reasoning** A verification approach that breaks complex systems into simpler subsystems, proves properties about each subsystem independently, and combines these proofs to establish system-wide guarantees. Critical for managing the complexity of large-scale autonomous systems.

**Cyber-Physical System** A system integrating computational elements with physical processes, where software controls physical actuators and sensors provide feedback. Drones are quintessential cyber-physical systems where incorrect computation can cause physical crashes.

**Depth-Semantic Gap** The disconnect between geometric perception (recognizing that a surface exists) and physical understanding (knowing whether that surface can support a drone's weight, has adequate friction, etc.). Current vision systems excel at geometry but lack physical semantics critical for safe contact.

**Digital Twin** A virtual replica of a physical system that synchronizes with real-world data to enable simulation, prediction, and testing. Digital Twins 2.0 refers to bidirectional systems that continuously update from real telemetry and can predict failures before they occur in physical hardware.

**Distributed Trust** Trust mechanisms that function without centralized authorities, enabling entities to verify each other's integrity and authenticate communications across organizational boundaries. Required for large-scale drone operations where no single authority controls all participants.

**Drone (Unmanned Aerial Vehicle/System - UAV/UAS)** An aircraft without a human pilot aboard that uses onboard computers to manage flight dynamics and execute flight commands delivered wirelessly. This report focuses primarily on small to mid-sized drones, operating across diverse missions including delivery, inspection, surveillance, agriculture, and emergency response. Also called Unmanned Aerial Vehicles (UAV) or Unmanned Aerial Systems (UAS).

**Dynamic Voltage and Frequency Scaling (DVFS)** A power management technique that adjusts processor voltage and clock speed to reduce energy consumption. In drones, DVFS can induce latency jitter that destabilizes flight control precisely when battery depletion makes energy management most critical.

**Edge Computing** Computational processing performed near the data source (e.g., on cell towers, base stations, or mothership platforms) rather than in distant cloud datacenters. Reduces latency for time-critical drone operations while managing bandwidth constraints.

**Elusive Constraints** Limitations or requirements that AI systems respect during training but violate in novel operational scenarios, only becoming apparent after deployment. It is a critical failure mode where neural networks behave correctly in known conditions but fail catastrophically in edge cases.

**Emergent Behavior** System-level behaviors that arise from interactions between components but cannot be predicted from analyzing individual components in isolation. In drone fleets, coordination algorithms may produce oscillations or deadlocks despite each vehicle's controller being individually stable.

**Federated Learning** A machine learning approach where multiple drones train models on local data and share only model updates (not raw data) with others, enabling fleet-wide learning while preserving privacy and reducing communication bandwidth.

**Fleet** A coordinated group of AI-powered drones that share data & experiences to achieve shared mission objectives and avoid catastrophic interference. Unlike swarms, each drone within a fleet may adopt specific, heterogeneous roles defined by the mission. A fleet can comprise drones with heterogeneous capabilities and limitations. Fleet learning allows each member to incrementally benefit from shared data.

**Formal Methods** Mathematical techniques for specifying, developing, and verifying software and hardware systems. They provide rigorous proofs of correctness but face scalability challenges with multi-million parameter AI models, requiring advances in compositional and probabilistic verification.

**FPGA (Field-Programmable Gate Array)** Reconfigurable hardware that can be programmed to perform specific computations with low latency and high energy efficiency. Offers advantages for real-time drone control but remains underutilized due to toolchain complexity and limited integration with AI frameworks.

**Geofencing** Virtual boundaries defined by GPS coordinates that restrict where drones can fly. Used for safety (preventing flight near airports) and security (protecting critical infrastructure), but vulnerable to GPS spoofing attacks.

**GPS Spoofing** An attack where false GPS signals deceive a drone about its location, potentially causing crashes or unauthorized access to restricted areas. Requires detection mechanisms and alternative navigation methods (visual odometry, ISAC positioning).

**Graceful Degradation** The ability of a system to maintain partial functionality when components fail or resources become constrained, rather than failing catastrophically. Critical for drones where sensor failures, communication loss, or battery depletion are inevitable.

**Graded Autonomy** A framework allowing smooth transitions between different levels of autonomy (fully autonomous, supervised, teleoperated) without destabilizing system behavior. Enables human operators to re-enter control loops during emergencies.

**Human-on-the-Loop** A supervisory control paradigm where humans monitor autonomous operations and intervene only during exceptions, as opposed to "human-in-the-loop" where humans actively control each action. Essential for scaling to large fleets where one operator supervises many vehicles.

**Hybrid AI Architecture** Systems combining learning-based components (neural networks for perception) with physics-based models (aerodynamic equations, control theory) to leverage both data-driven flexibility and principled guarantees about physical behavior.

**ISAC (Integrated Sensing and Communication)** A 6G capability where the same radio waveform serves both communication (data transmission) and sensing (radar, positioning) functions simultaneously. Offers spectrum efficiency but creates security vulnerabilities where sensing and communication cannot be decoupled.

**Jamming** Deliberate interference with communication or navigation signals to disrupt drone operations. Counter-jamming techniques include frequency hopping, directional antennas, and alternative navigation methods.

**JCAS (Joint Communication and Sensing)** See ISAC. Technologies enabling centimeter-level positioning accuracy through the integration of communication and sensing functions, critical for GPS-denied indoor/outdoor navigation.

**LLM (Large Language Model)** AI models like GPT trained on massive text datasets, capable of understanding and generating human language. In drone contexts, enables natural language mission planning ("deliver to north entrance if clear, otherwise wait") and agentic systems that reason about tasks autonomously.

**Model-Based Systems Engineering (MBSE)** A formalized approach to systems engineering that uses explicit models to capture requirements, design, analysis, and verification rather than relying solely on documents. MBSE for AI extends this to represent learning processes, uncertainty bounds, and training data provenance.

**Mothership Architecture** A fleet coordination paradigm where a larger platform (aerial or ground-based) serves as mobile edge infrastructure, providing drones with task allocation, situational awareness, automated docking/recharging, and persistent mission management.

**Multi-Agent System** A system composed of multiple autonomous entities (agents) that interact to achieve individual or collective goals. Drone fleets are multi-agent systems where coordination emerges from local decisions and communication.

**Non-Deterministic** Systems or algorithms that may produce different outputs for identical inputs depending on internal state, randomness, or environmental factors. Deep learning models are inherently non-deterministic, breaking traditional verification approaches that assume reproducibility.

**Over-the-Air (OTA) Update** Remote software updates delivered wirelessly to deployed systems. For drone fleets, OTA updates must preserve timing contracts and safety guarantees while enabling rapid deployment of improved AI models across thousands of vehicles.

**Part 107** The FAA regulation (14 CFR Part 107) governing commercial small unmanned aircraft operations in the United States. Requires Remote Pilot Certification but addresses piloted operations, not the autonomous fleet challenges this report focuses on.

**Physics-Aware AI** AI systems that incorporate physical principles (aerodynamics, gravity, inertia) into their architectures, ensuring predictions and decisions respect physical constraints rather than purely statistical patterns. Critical for safe physical interaction.

**Probabilistic Model Checking** A verification technique that computes the probability of a system satisfying certain properties, providing bounds on the likelihood of unsafe behaviors rather than absolute guarantees. Useful for systems with inherent uncertainty like learning-based autonomy.

**Real-Time System** A system where correctness depends not only on logical results but on meeting timing deadlines. Flight control is a hard real-time system where missing a 10ms deadline can cause crashes, unlike soft real-time systems (video streaming) where occasional delays are tolerable.

**Runtime Verification** Monitoring deployed systems during operation to detect violations of safety properties, enabling interventions before failures occur. Complements pre-deployment verification for systems that evolve or operate in unpredictable environments.

**Sim-to-Real Gap** The discrepancy between simulated and physical environments that causes AI models trained in simulation to fail when deployed on real hardware. Arises from unmodeled dynamics (wind, sensor noise) and subtle environmental factors (lighting, textures).

**Swarm** A tightly coupled multi-agent system where many simple agents coordinate through local interactions to achieve collective behaviors (flocking, pattern formation). Distinct from fleets, which may be heterogeneous and structured.

**Testbed** A controlled environment for experimenting with and validating systems before operational deployment. A national drone testbed would provide shared infrastructure (flight ranges, edge computing, communication networks) and datasets accessible to researchers nationwide.

**UAS-CTI (Unmanned Aircraft Systems Collegiate Training Initiative)** An FAA program recognizing universities, colleges, and technical schools that prepare students for careers in drone operations and technology. Currently focuses primarily on piloting rather than computing challenges.

**Value-of-Information (VoI)** A decision-making framework that allocates resources (sensing, computation, communication, energy) based on how much information gathering reduces mission uncertainty or risk, rather than using fixed policies. Enables adaptive resource management under constraints.

**WCET (Worst-Case Execution Time)** The maximum time a computational task can take under any possible conditions (worst-case sensor data, maximum memory usage, etc.). Critical for real-time systems where safety depends on guaranteeing that control loops complete within bounded time.

# 6. Appendix

## 6.1 Workshop Structure

Workshop Agenda

December 1-2, 2025
The Darcy Hotel
1515 Rhode Island Ave NW, Washington, DC 20005

**December 1, 2025 (Monday)**

| Time | Session Description | Room |
|---|---|---|
| 8:30 am | Breakfast Available | Ellington |
| 8:30 am | Registration | Logan Ballroom Foyer |
| 9:30 am | Welcome and Overview of the Workshop | Bader Ginsberg |
| 10:00 am | Lightning Introductions Part 1 | Bader Ginsberg |
| 10:20 am | Coffee Break and Brainstorming Activity | Bader Ginsberg |
| 10:40 am | **Speakers:**<br>● Guoquan "Paul" Huang, University of Delaware<br>● Andy Thurling, DroneUp | Bader Ginsberg |
| 11:30 am | Lightning Introductions Part 2 | Bader Ginsberg |
| 12:00 pm | Introduce breakout group activity and Group Photo | Bader Ginsberg |
| 12:30 pm | Lunch | Gerrard St. Kitchen<br>(Hotel Restaurant) |
| 1:30 pm | **Breakout Groups Round 1**<br>● Current Challenges and Opportunities | Logan East, Logan West, Bader Ginsberg |
| 3:00 pm | Full Group Report Out and Discussion | Bader Ginsberg |
| 3:30 pm | Break | Logan Ballroom Foyer |
| 4:00 pm | **Speaker:** | Bader Ginsberg |

| | • Sabine Brunswicker, Purdue University | |
|---|---|---|
| 4:30 pm | **Breakout Groups Round 2**<br>• Dreaming of 2035 | Logan East, Logan West, Bader Ginsberg |
| 5:45 pm | Full Group Report Out and Discussion | Bader Ginsberg |
| 6:30 pm | Dinner | Ellington |

**December 2, 2025 (Tuesday)**

| **Time** | **Session Description** | **Room** |
|---|---|---|
| 8:00 am | Breakfast Available | Ellington |
| 9:00 am | Welcome to Day 2 | Bader Ginsberg |
| 9:10 am | **Speaker:**<br>• Matt Welsh | Bader Ginsberg |
| 9:30 am | **Breakout Groups Round 3**<br>• What is missing from this conversation? | Logan East, Logan West, Bader Ginsberg |
| 11:00 am | Break | Logan Ballroom Foyer |
| 11:30 am | Full Group Report Out and Discussion | Bader Ginsberg |
| 12:30 pm | Lunch | Gerrard St. Kitchen (Hotel Restaurant) |
| 1:30 pm | **Breakout Groups Round 4**<br>• Synthesis and Recommendations | Logan East, Logan West, Bader Ginsberg |
| 2:50 pm | Wrap Up and Closing Remarks | Bader Ginsberg |

## 6.2 Workshop Attendees

| First Name | Last Name | Affiliation |
|---|---|---|
| Nils | Aschenbruck | Osnabrück University |
| Sabine | Brunswicker | Purdue University |
| Kevin | Butler | University of Florida |
| Dilma | Da Silva | National Science Foundation (NSF) |
| Karthik | Dantu | University at Buffalo |
| Sajal | Das | Missouri University of Science and Technology |
| Warren | Dixon | University of Florida |
| Sebastian | Elbaum | University of Virginia |
| Kelvin (Kun) | Fu | University of Delaware |
| Ajay | Gupta | Western Michigan University |
| Guoquan (Paul) | Huang | University of Delaware |
| Anne Marie | Kelly | IEEE-CS |
| Mike | Kronander | Burgess & Niple, Inc. |
| Giuseppe | Loianno | University of California, Berkeley |
| Antonio | Loquercio | University of Pennsylvania |
| Mark | Mueller | University of California, Berkeley |
| Terry | Mullins | Burgess & Niple, Inc. |
| Kamesh | Namuduri | University of North Texas |
| Adam | Norton | University of Massachusetts Lowell |
| Dario | Pampili | Rutgers University |
| Nikolaos | Papanikolopoulos | University of Minnesota |
| Srini | Parthasarathy | Ohio State University |
| Lakshmish | Ramaswamy | University of Georgia |
| Aanjhan | Ranganathan | Northeastern University |
| Shashi | Shekhar | University of Minnesota |
| Weisong | Shi | University of Delaware |
| Yiyu | Shi | University of Notre Dame |
| Deborah | Silver | Rutgers University |
| Christopher | Stewart | Ohio State University |
| Maxwell | Taylor | University of Montana/Idaho National Labs |
| Andy | Thurling | DroneUp |
| Ufuk | Topcu | University of Texas |
| Damla | Turgut | University of Central Florida |
| Selcuk | Uluagac | National Science Foundation (NSF) |
| Peng | Wei | George Washington University |
| Matt | Welsh | Palantir |
| Dongyan | Xu | Purdue University |
| Holly | Yanco | University of Massachusetts Amherst |
| Hongwei | Zhang | Iowa State University |
| Ting | Zhu | The Ohio State University |

## 6.3 Acknowledgements

# 7. References

American Society of Civil Engineers. (2025, March). *A Comprehensive Assessment of America's Infrastructure.* https://infrastructurereportcard.org/wp-content/uploads/2025/03/Full-Report-2025-Natl-IRC-WEB.pdf

Department of Defense. (2024, August). *Structuring Change to Last: An Update on Innovation at the Department of Defense*. DOD Innovation Fact Sheet. https://media.defense.gov/2024/Aug/07/2003519333/-1/-1/0/DOD-INNOVATION-FACT-SHEET-AUGUST-2024.PDF

Federal Aviation Administration. (n.d.). *FAA Aerospace Forecast | Fiscal Years 2025-2045*. https://www.faa.gov/data_research/aviation/aerospace_forecasts/FY-2025-2045-Full-Forecast-Document-and-Tables.pdf

Fortune Business Insights. (2026, April 13). *Drone Services Market Size, Share | Growth Report [2034]*. Drone Services Market Size, Share, By Service Type, By Application, By End-use Industry, and Regional Forecast, 2026-2034. Retrieved April 29 2026, from https://www.fortunebusinessinsights.com/drone-services-market-102682

Hashmi, Z. G., Griffin, R., Espinola, J. A., Sullivan, A. F., Boggs, K. M., Swanton, M., Jarman, M. P., Jansen, J. O., Kerby, J. D., & Camargo, C. A., Jr. (2026). Population Access to US Trauma Centers and Teletrauma-Using Emergency Departments. *JAMA network*, 9(2), e2556958. https://doi.org/10.1001/jamanetworkopen.2025.56958

HAPS Alliance. (2026, March). *6G from the Stratosphere The Role of HAPS in Key Use Cases*. https://hapsalliance.org/wp-content/uploads/formidable/12/HAPSAlliance_6G_WhitePaper_MAR26.pdf

IEEE Standards Association. (2022). IEEE Trial-Use Standard for Aerial Network Communication. *IEEE Std 1920.1-2022*. http://doi.org/10.1109/IEEESTD.2023.10120679

Lightcast. (2026). Lightcast Data. https://lightcast.io

National Oceanic and Atmospheric Administration. (n.d.). *Billion-dollar weather and climate disasters*. Billion-Dollar Weather and Climate Disasters | National Centers for Environmental Information. Retrieved April 28 2026, from https://www.ncei.noaa.gov/access/billions/

Odegard, K. (2025, January 17). *With farms in slump, U.S. tractor sales slow down in 2024.* Capital Press.

https://capitalpress.com/2025/01/17/with-farms-in-slump-u-s-tractor-sales-slow-down-in-2024/

Peters, A. (2026, April 13). *These medical delivery drones could soon be supplying U.S. hospitals.* Fast Company. https://www.fastcompany.com/40555697/zipline-will-soon-be-flying-its-lifesaving-drones-in-the-u-s

Skydio. (n.d.) *When lives are on the line, get Skydio in the air.* Drone as First Responder (DFR) for Public Safety. Retrieved May 1 2026, from https://www.skydio.com/solutions/dfr

United States Forest Service. (2024, January). *2024 UAS Annual Report.* US Forest Service. https://www.fs.usda.gov/sites/default/files/FY24-FS-UAS-Report.pdf

United States Government Accountability Office. (2023, April 19). *Aviation workforce: Supply of airline pilots and aircraft mechanics* (GAO-23-106769). https://www.gao.gov/products/gao-23-106769